\documentclass[final]{elsarticle}
\usepackage{hyperref}

\usepackage[T1]{fontenc}
\usepackage{graphicx}
\usepackage{bm}
\usepackage{mathrsfs}
\usepackage{amsmath}
\usepackage{amssymb}
\usepackage{algorithm}
\usepackage{algorithmic}
\usepackage{array}
\usepackage{bm}
\usepackage{booktabs}
\usepackage{caption3}
\usepackage{url}
\usepackage{multirow}
\usepackage{xcolor}
\usepackage{lscape}
\newcommand{\myrevisedcolor}[1]{\textcolor{black}{{}#1}}

\journal{Journal of \LaTeX\ Templates}

\begin{document}
\sloppy
\begin{frontmatter}

\title{An Accelerated Correlation Filter Tracker}

\author[jiangnanaddress,surreyaddress]{Tianyang Xu}

\author[surreyaddress]{Zhen-Hua Feng}

\author[jiangnanaddress]{Xiao-Jun Wu\corref{corrtext}}
\cortext[corrtext]{Corresponding author} \ead{e-mail:wu\_xiaojun@jiangnan.edu.cn}

\author[surreyaddress]{Josef Kittler}

\address[jiangnanaddress]{School of Internet of Things Engineering, Jiangnan University, Wuxi 214122, China}
\address[surreyaddress]{Centre for Vision, Speech and Signal Processing, University of Surrey, Guildford, GU2 7XH, UK}

\begin{abstract}
Recent visual object tracking methods have witnessed a continuous improvement in the state-of-the-art with the development of efficient discriminative correlation filters (DCF) and robust deep neural network features. Despite the outstanding performance achieved by the above combination, existing advanced trackers suffer from the burden of high computational complexity of the deep feature extraction and online model learning. We propose an accelerated ADMM optimisation method obtained by adding a momentum to the optimisation sequence iterates, and by relaxing the impact of the error between DCF parameters and their norm. The proposed optimisation method is applied to an innovative formulation of the DCF design, which seeks the most discriminative spatially regularised feature channels. A further speed up is achieved by an adaptive initialisation of the filter optimisation process. The significantly increased convergence of the DCF filter is demonstrated by establishing the optimisation process equivalence with a continuous dynamical system for which the convergence  properties can readily be derived. The experimental results obtained on several well-known benchmarking datasets demonstrate the efficiency and robustness of the proposed ACFT method, with a tracking accuracy comparable to the start-of-the-art trackers.
\end{abstract}


\end{frontmatter}


\section{Introduction}
Visual object tracking is a fundamental research topic in computer vision and artificial intelligence, with undiminishing practical demands emanating from video surveillance, medical image processing, human-computer interaction, robotics and so forth. 
A tracking system aims at locating a target in a video sequence given its initial state. 
Over the decades of development, numerous algorithms have been proposed for designing robust trackers with a certain degree of success~\cite{zhou2016graph,wang2017deep,dong2018hyperparameter}.
As a byproduct, a variety of experimental protocols and evaluation metrics have been proposed to support academic research in terms of standard  benchmarks~\cite{Wu2015Object} and technology pushing challenges~\cite{kristan2018sixth,kristan2019seventh,xu2016fast,du2019visdrone}. 
However, there remain many challenging factors, such as background clutter, non-rigid deformation, abrupt motion, occlusion and real-time requirement, that impede accurate tracking performance in unconstrained scenarios.

Among the recently advanced trackers, a notable shift of focus of the research effort has been towards exploiting the notion of discriminative correlation filter (DCF) and deep convolutional neural network (CNN) representations~\cite{li2018deep}. 
DCF-based trackers efficiently augment original image samples via circulant shift and consider the training process as ridge regression~\cite{Henriques2012Exploiting}. 
The entire set of circulant shifts satisfies the convolution theorem, simplifying the computation via element-wise multiplication and division in the frequency space, rather than calculating the product or inverse of training matrices. 
These advantages are further strengthened by incorporating additional regularisations and constraints~\cite{Danelljan2015Learning,xu2018learning}.
Apart from the improvements achieved by sophisticated mathematical formulations, advanced DCF trackers tend to employ powerful deep CNN features for boosting the performance~\cite{dong2016occlusion,wang2018semi,wang2017video}.
The trackers equipped with robust deep CNN features have outperformed those with traditional hand-crafted features in recent VOT challenges~\cite{kristan2018sixth}, while feature selection has been demonstrated as one of the most essential mechanisms enabling improved tracking performance~\cite{wang2015understanding,xu2019joint}. 
Despite the recent success with graphics processing unit (GPU) implementations, it is time-consuming to extract deep CNN features and to learn complex appearance models involving a high-volume of variables in the DCF formulation on-line.

To mitigate the above issues, we propose a method for constructing an accelerated correlation filter tracker (ACFT) by incorporating a momentum in the gradient based updates of the filter. We also relax the impact of the discrepancy between the filter parameters driven by data on one hand, and a regularisation of their norm on the other. We determine the convergence properties of the accelerated and relaxed ADMM algorithm~\cite{francca2018relax} by establishing its link to a continuous dynamic system by finding the continuous limit of the sequence of its iterates. 
The connection between iterative optimisation and continuous dynamical systems has been widely studied recently, leading to improved algorithms~\cite{wibisono2016variational,francca2018admm} and enhanced  understanding of their stability and convergence properties. 
The algorithm convergence is further sped up by means of an adaptive initialisation of the iterative optimisation process.
Accordingly, in our ACFT, the filter optimisation process for frame $k$ is initialised by the filter parameters found at frame $k-1$. In fact the proposed adaptive filter initialisation process imposes a temporal smoothness constraint on the solution, thus achieving better accuracy in tracking.

To avoid the computational complexity of extracting CNN features using the latest deep CNNs, we opt for the AlexNet~\cite{krizhevsky2012imagenet}, in tandem with hand-crafted features, \textit{i.e.} histogram of oriented gradients (HOG) and Colour Names (CN).
More powerful deep CNN features, such as ResNet~\cite{he2016deep}, are also analysed in our experiments to demonstrate the merit of R\_A-ADMM in the regularised DCF paradigm over the traditional ADMM.
The results of extensive experiments obtained on several well-known visual object tracking benchmarks,~\textit{i.e.}, OTB2013, TB50, OTB2015 and VOT2017, demonstrate the efficiency and robustness of the proposed ACFT method.

The main innovations of ACFT include:
\begin{itemize}
\item A formulation of the DCF design problem which focuses on informative feature channels and spatial structures by means of novel  regularisation. 
\item A proposed relaxed optimisation algorithm referred to as R\_A-ADMM for optimising the regularised DCF. In contrast to the standard ADMM, the algorithm achieves a convergence rate of $\mathcal{O}(1/l^2)$ instead of ($\mathcal{O}(1/l)$) for ADMM.
\item A novel temporal smoothness constraint, implemented by an adaptive initialisation mechanism, to achieve further speed up via transfer learning among video frames.
\item The proposed adoption of AlexNet to construct a light-weight deep representation with a tracking accuracy comparable to more complicated deep networks, such as VGG and ResNet.
\item An extensive evaluation of the proposed methodology on several well-known visual object tracking datasets, with the results confirming the acceleration gains for the regularised DCF paradigm.
\end{itemize}

The paper is structured as follows. 
In section~\ref{relatedwork}, we discuss the prior work on related theories of DCF and optimisation methods.
We formulate our tracking task and optimisation approach in Section~\ref{proposed}, and study its properties.
The experimental results are reported and analysed in Section~\ref{experiment}.
The conclusions are  drawn  in Section~\ref{conclusion}.

\section{Related Work}\label{relatedwork}
The continuous improvement in visual object tracking recorded over the last two decades stem from two main factors: i) advances in the tracking problem formulation, and ii) improved efficiency of numerical optimisation techniques used to solve the tracking problem.
In this section, we briefly review the pertinent theories and approaches in visual tracking, focusing on DCF, and associated iterative optimisation methods. 
For more detailed introductions to visual object tracking, readers are referred to recent survey papers~\cite{Kristan2017a}.

\subsection{DCF tracking formulations}
Normalised cross correlation (NCC)~\cite{Briechle2001Template} is the seminal work that introduced cross correlation as an efficient similarity metric into template matching.
This technique was further improved by the Minimum Output Sum of Squared Error (MOSSE)~\cite{Bolme2010Visual} filter.
MOSSE was proposed to achieve fast object tracking via correlation, with the objective of minimising the sum of squared errors.
The cornerstone of DCF was laid by Henriques,~\textit{et al.}~\cite{Henriques2012Exploiting}.
In DCF, the tracking task is achieved in the tracking-by-detection fashion. The DCF design exploits circulant structure~\cite{Gray2006Toeplitz} of the target data represented  with kernels (CSK)~\cite{Henriques2012Exploiting}.
In CSK, augmented training samples generated by the circulant structure are efficiently calculated in the frequency domain.
The Fast Fourier Transform (FFT) employed for mapping variables across the spatial and spatial frequency domains enables further acceleration of the computation.
The computational efficacy of DCF received a wide attention, stimulating improved formulations in multi-response fusion (Staple)~\cite{Bertinetto2016Staple}, circulant sparse representation (CST)~\cite{Zhang2016In}, support vector filters (SCF)~\cite{Zuo2016learning}, and so on.

The original DCF formulation via ridge regression provides efficient optimisation in the frequency domain. However, it lacks the ability to select a salient spatial region to support discriminative filtering.
Specifically, the DCF technique learns filters of the same size as the input features corresponding to the entire search window.
While, a larger search window brings discriminative information from the background, it introduces additional spatial distortion from circulant samples.
To alleviate the growing spatial distortion problem, Danelljan,~\textit{et al.}, proposed to learn spatially regularised correlation filters (SRDCF)~\cite{Danelljan2015Learning}.
A pre-defined weighting window is employed to force the energy of the filters to concentrate in the central region of the search window, decreasing the number of distorted samples.
A similar idea is developed in the background-aware correlation filter (BACF)~\cite{Galoogahi2017Learning} and the spatial reliability strategy (CSRDCF)~\cite{Lukezic2017Discriminative}.
The difference is that BACF utilises spatial sample pruning to reduce the boundary effect and CSRDCF exploits a colour histogram based foreground-background mask to filter out non-target region.

Besides the spatial appearance modelling techniques, temporal appearance clues have also been explored in the DCF paradigm.
To alleviate tracking shifts and failures, the long-term correlation tracker (LCT)~\cite{Ma2015Long} was proposed to train an online random fern classifier to re-detect objects under certain pre-defined conditions.
Adaptive decontamination of the training set (SRDCFdecon)~\cite{danelljan2016adaptive} was designed to make more historical frames available for the filter learning, dynamically managing the training set by estimating the quality of the temporal samples, with less weights for the non-essential appearance. 
The subsequent continuous convolution operators (C-COT) improved the decontamination of the appearance model by learning continuous representations, achieving sub-grid tracking accuracy.
However, it is computationally demanding for SRDCFdecon and C-COT to involve a large bundle of frames in the optimisation, resulting in poor tracking speed.
To resolve the conflict between simultaneous temporal clues collection and efficient optimisation, efficient convolution operator (ECO)~\cite{Danelljan2016ECO} was proposed to cluster historical frames based on their similarities.
In addition, dimension reduction is employed by ECO to further decrease the feature channels, especially for multi-channel CNN representations.
Conversely, the spatial-temporal regularised correlation filter (STRCF)~\cite{li2018learning} was proposed to combine temporal smoothness and spatial filtering together, realising a simple, but effective method to take temporal appearance into account.

With the development of the above formulations, the recently proposed robust DCF trackers have achieved outstanding performance~\cite{xu2019learningL,lu2018deep,lu2019adaptive}, especially when equipped with powerful deep CNN features. However, in the drive to improve the tracking performance, the computational complexity of the algorithms has been largely overlooked. 
In particular, little attention has been focused on accelerating DCF formulations exploiting deep features.
To balance the tracking accuracy and efficiency, one of the aims of this paper is to analyse different deep CNN representations and propose a method of constructing a light-weight DCF tracker.

\subsection{Iterative optimisation methods}
The original DCF formulation aims at solving a ridge regression problem, which can be efficiently obtained with a closed-form solution in the frequency domain.
However the solution is more complicated in the case of the regularised DCF formulations,~\textit{e.g.}, SRDCF, CSRDCF and C-COT, that have to perform iterative optimisation due to their complex regularisation terms.
Generally, SRDCF and C-COT consider their regularised objectives as 
large linear systems.
Classical mathematical methods,~\textit{e.g.}, Gauss-Seidel, Conjugate Gradient, are commonly employed.
More recently, ADMM attracted considerable attention as a technique for solving regularised DCF problems, such as in BACF, CSRDCF and STRCF.
Compared with solving large linear systems, ADMM is more intuitive in solving an objective with multiple regularisation terms~\cite{xu2018non}. 

Recently, multiple developments in the field of optimisation have advanced the state of the art of the discipline considerably. First of all, the speed of convergence of iterative gradient based optimisation algorithms has been dramatically improved by the Nestorov's method of taking the momentum of variable updates into consideration. This idea is applicable to the ADMM type algorithms of interest in this paper. Second, in the case of multi-objective criterion functions, the forces exerted by the different components of its gradient in an ADMM algorithm are difficult to reconcile, leading to an oscillatory behaviour, and therefore a slow convergence. This problem can be mitigated by the idea of a relaxed ADMM, which enables the contribution of any discrepancy between target solutions to be controlled parametrically. The third important step forward has been achieved by the establishment of the connection between iterative, gradient based optimisation algorithms and  continuous dynamical systems. The connection can be determined by taking the sequence of iterates of an algorithm to a continuous limit~\cite{su2016differential}. The connection has a very important implication. It allows to study the convergence rate and stability of iterative optimisation algorithms by reference to the Lyapunov theory applied to the equivalent dynamic systems. 
In other words, the process of gradient-based optimisation can be considered as solving a continuous dynamical system~\cite{su2016differential}. The connection can also be exploited by designing new gradient-based optimisation algorithms via differential equations in the continuous domain~\cite{krichene2015accelerated,wilson2016lyapunov}. 
As for the ADMM family,~\textit{e.g.}, relaxed ADMM (R\_ADMM), relaxed and accelerated ADMM (R\_A-ADMM) and relaxed heavy ball ADMM, their properties  have been studied via differential equations obtained by the means of their continuous limit in \cite{francca2018admm,francca2018relax}.

In this paper we apply these innovations in  optimisation in our unique, dimensionality reducing formulation of the DCF problem. In the adopted ADMM optimisation framework, we relax the discrepancy between the target solution for our DCF, driven by the circulant matrix data, and the requirement of its minimum norm. We also add a parameterised momentum to the variable updates to achieve a convergence speed up. Using the continuous limit analysis of the proposed iterative optimisation process we derive the convergence rate of our accelerated and relaxed ADMM algorithm as a function of the relaxation parameter and a damping factor. The dramatic algorithmic speed up is enhanced by the proposed light weight deep neural network architecture developed for realising the discriminative correlation filter.     
A further acceleration of the filter design is achieved by an adaptive initialisation of the iterative optimisation process, reflecting a smooth temporal modelling of the DCF evolution.

\begin{figure}[!t]
\centering
\includegraphics[trim={0mm 90mm 35mm 0mm},clip,width=1\linewidth]{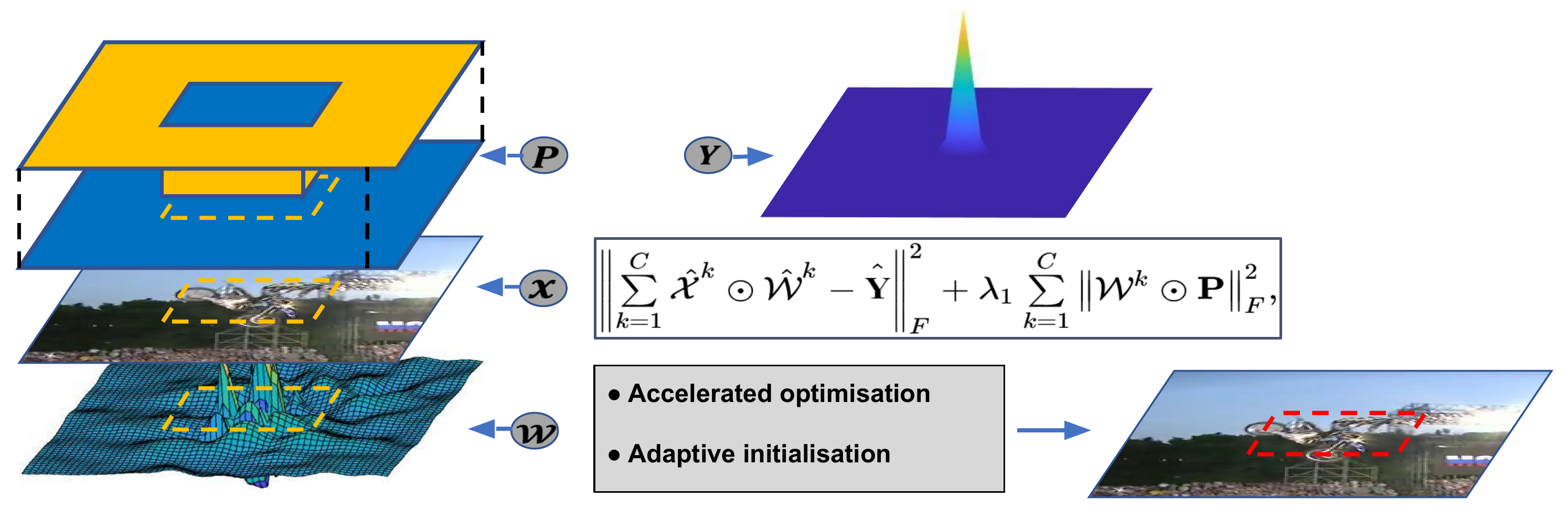}
\caption{We propose to use R\_A-ADMM for efficient optimisation of our mask-constrained learning objective. The multi-feature combination is considered in our implementation to improve the robustness and accuracy.}
\label{overview}
\end{figure}

\section{The Proposed Approach}
\label{proposed}
\subsection{Objective}
In the filter learning stage, a square image patch $\mathbf{S}$ ($N_S\times N_S$ pixels) centred at the target is cropped as a search window, with a predefined parameter $\beta$ controlling the area ratio between the search window and the target.
We then extract feature representations from $\mathbf{S}$, using HOG, CN and AlexNet~\cite{krizhevsky2012imagenet} features.
Note that HOG and CN are hand-crafted features to enhance representation of the visual input in terms of edge and colour information respectively.
We employ the Relu output of Conv-4 and Conv-5 block from AlexNet as deep features.
Although more recent networks, \textit{e.g.}, ResNet~\cite{he2016deep}, can provide more powerful feature representations, they are much more demanding in terms of computer memory and computational complexity and have been deliberately avoided to speed up the filter design process.
We denote the feature representations as $\mathcal{X}\in\mathbb{R}^{N\times N\times C}$, where $C$ is the channel dimension and $N\times N$ is the spatial resolution with the corresponding $N_S/N$ stride (stride $4$ for hand-crafted features and $16$ for deep features).

We aim to learn a system of discriminative filters with parameters $\mathcal{W}\in\mathbb{R}^{N\times N\times C}$ that regress the input $\mathcal{X}$ to a predefined desired label matrix $\mathbf{Y}\in\mathbb{R}^{N\times N}$. 
The learning objective is formulated as follows:
\begin{equation}
\label{obj}
\mathcal{W} = \arg\underset{\mathcal{W}}{\min}
\left\|\sum\limits_{k=1}^{C}\mathcal{X}^k\circledast\mathcal{W}^k-\textbf{Y}\right\|_F^2+ \lambda_1\sum\limits_{k=1}^{C}\left\|\mathcal{W}^k\right\|^2_F,
\end{equation}
where $\circledast$ is the circular correlation operator (cyclic correlation)~\cite{Henriques2015High}, $||*||_F$ is the Frobenius norm of a matrix, $\mathcal{X}^k$ and $\mathcal{W}^k$ denote the $k$-th channel of $\mathcal{X}$ and $\mathcal{W}$, and $\lambda_1$ is a regularisation parameter.

It is generally known that the discrimination of DCF can be enhanced using a larger $\beta$ in order to cover more background information in the search window.
Circular correlation is equivalent to extending the border by a circulant padding. However, this leads to spatial distortion.
To mitigate this problem, spatial regularisation is introduced in the DCF paradigm via different techniques~\cite{Danelljan2015Learning,Lukezic2017Discriminative,Galoogahi2017Learning,xu2018learning} to concentrate the filter energy in the centre region. As a result of this measure, the validity of the region encompassed by circular correlation increases, even with a large $\beta$. 
In addition, a prior \textit{cosine} weight window~\cite{Henriques2015High} is applied to the search region in order  to suppress the context near the search region border and thus to alleviate the spatial distortion.
We follow the basic idea of spatial regularisation and constrain the elements outside the centre to zero: $
w_{i,j}^k=0,\ \textrm{if}\ (i,j)\in\mathcal{B},
$ 
where $w_{i,j}^k$ is the element of the $i$-th row and $j$-th column of the $k$-th channel in $\mathcal{W}$, and $\mathcal{B}$ denotes the spatial region corresponding to the background.

As shown in Fig.~\ref{overview}, to combine the constraint into our objective, we introduce another regularisation term using an indicator matrix $\mathbf{B}\in\mathbb{R}^{N\times N}$ with zeros for the target region and ones for the background region, $\lambda_2\left\|\mathcal{W}^k\odot\mathbf{B}\right\|^2_F$, which can further be fused with the existing regularisation term $\lambda_1\left\|\mathcal{W}^k\right\|^2_F$ for each channel:
\begin{equation}
\label{fuse}
\lambda_1\left\|\mathcal{W}^k\right\|_F^2+\lambda_2\left\|\mathcal{W}^k\odot\mathbf{B}\right\|_F^2 = \lambda_1\left\|\mathcal{W}^k\odot\mathbf{P}\right\|^2_F
\end{equation}
where $\mathbf{P}=\sqrt{\mathbf{1}+\frac{\lambda_2}{\lambda_1}}\mathbf{B}$, $\mathbf{1}$ is an all-ones matrix of the same size as $\mathbf{B}$, and $\odot$ denotes element-wise multiplication.
The learning objective can now be reformulated to minimise the following function:
\begin{equation}
\label{objre}
\Lambda\left(\mathcal{W}\right)=\left\|\sum\limits_{k=1}^{C}\hat{\mathcal{X}}^k\odot\hat{\mathcal{W}}^k-\hat{\textbf{Y}}\right\|_F^2+ \lambda_1\sum\limits_{k=1}^{C}\left\|\mathcal{W}^k\odot\mathbf{P}\right\|^2_F,
\end{equation}
where $\hat{\cdot}$ denotes the Fourier representation~\cite{Henriques2012Exploiting}. 

According to Plancherel's theorem, the circular correlation in the spatial domain is equivalent to element-wise multiplication in the frequency domain.
To achieve efficient optimisation across the domains, we consider the Fourier representation as an auxiliary variable $\hat{\mathcal{U}}^k=\mathcal{F}\left(\mathcal{W}^k\right)$, where $\mathcal{F}$ and $\mathcal{F}^{-1}$ denotes forward and inverse Fourier transform, respectively. 
As $\mathcal{F}$ is a full row/column rank linear transform, the objective can be naturally considered as augmented Lagrangian:
\begin{equation}
\label{aug}
\mathcal{L}_{\rho}\left(\hat{\mathcal{U}},\mathcal{W},\mathcal{T}\right)=f\left(\hat{\mathcal{U}}\right)+g\left(\mathcal{W}\right)+h_{\rho}\left(\hat{\mathcal{U}},\mathcal{W},\mathcal{T}\right),
\end{equation}
\begin{equation}\label{aug_d}
\left\{
\begin{aligned}
&f\left(\hat{\mathcal{U}}\right)=\left\|\sum\limits_{k=1}^{C}\hat{\mathcal{X}}^k\odot\hat{\mathcal{U}}^k-\hat{\textbf{Y}}\right\|_F^2\\
&g\left(\mathcal{W}\right)=\lambda_1\sum\limits_{k=1}^{C}\left\|\mathcal{W}^k\odot\mathbf{P}\right\|^2_F\\
&h_{\rho}\left(\hat{\mathcal{U}},\mathcal{W},\mathcal{T}\right)=\sum\limits_{k=1}^{C}\frac{\rho}{2}\left\|\mathcal{F}^{-1}\left(\hat{\mathcal{U}}^k\right)-\mathcal{W}^k+\mathcal{T}^k\right\|^2_F\\
\end{aligned}
\right.
\end{equation}
where $\mathcal{T}$ is the Lagrange multiplier and $\rho$ is the penalty.

\begin{algorithm}[t]
\begin{algorithmic}
\vspace{0.03in}
\STATE \textbf{Input:} $f\left(\hat{\mathcal{U}}\right)$, $g\left(\mathcal{W}\right)$, $h_{\rho}\left(\hat{\mathcal{U}},\mathcal{W},\mathcal{T}\right)$, penalty $\rho$ and $\alpha$ 
\STATE \textbf{initialise} $l=0$, $\hat{\mathcal{U}}[0]$, $\mathcal{W}[0]$ and $\mathcal{T}[0]$
\STATE \textbf{while} convergence condition \textbf{not} satisfied, \textbf{do}
\STATE \ \ \ \ 1. $\hat{\mathcal{U}}[l+1]\leftarrow\arg\underset{\hat{\mathcal{U}}}{\min}\left\{f\left(\hat{\mathcal{U}}\right)+h_{\rho}\left(\hat{\mathcal{U}},\mathcal{W}[l],\mathcal{T}[l]\right)\right\}$
\STATE \ \ \ \ 2. $\hat{\mathcal{V}}\leftarrow\alpha\hat{\mathcal{U}}[l+1]+\left(1-\alpha\right)\mathcal{F}\left(\mathcal{W}[l]\right)$
\STATE \ \ \ \ 3. $\mathcal{W}[l+1]\leftarrow\arg\underset{\mathcal{W}}{\min}\left\{g\left(\mathcal{W}\right)+h_{\rho}\left(\hat{\mathcal{V}},\mathcal{W},\mathcal{T}[l]\right)\right\}$
\STATE \ \ \ \ 4. $\mathcal{T}[l+1]\leftarrow\mathcal{T}[l]+\mathcal{F}^{-1}(\hat{\mathcal{V}})-\mathcal{W}[l+1]$
\STATE \ \ \ \ 5. $l=l+1$
\STATE \textbf{end for}
\vspace{0.03in}
\end{algorithmic}
\caption{Optimisation via the R\_ADMM and ADMM for problem (\ref{aug}). The relaxation parameter $\alpha\in(0,1)\cup(1,2)$ for R\_ADMM and $\alpha=1$ for ADMM.}
\label{alg2}
\end{algorithm}

\subsection{Optimisation}
The basic DCF formulation~\cite{Bolme2010Visual,Henriques2012Exploiting} achieves superior efficiency thanks to the circulant matrix and Fourier transform. 
To take a full advantage of both, the discriminant filter learning must be conducted in the frequency domain.  
Note that only the basic $\ell_2$-norm regularisation term can be opted for to maintain a closed-form solution.
In contrast, more recent techniques shift their focus on effective clues in the original spatial domain~\cite{Lukezic2017Discriminative,Galoogahi2017Learning,Danelljan2016ECO}, which requires performing iterative optimisation in the two domains.
Though these advanced techniques have demonstrated their advantages in terms of tracking performance, this is achieved at the expense of increased computational complexity, especially for CNN feature extraction and iterative optimisation.

Following the standard iterative method in optimising multiply regularised DCF objectives~\cite{Galoogahi2017Learning,Lukezic2017Discriminative,li2018learning}, we present the corresponding ADMM and R\_ADMM optimisation steps for our objective in Equ.~\ref{aug} in Algorithm~\ref{alg2}. 
Compared to ADMM, R\_ADMM introduces an additional relaxation variable $\hat{\mathcal{V}}$ (in step 2) to perform a moving average between the current $\hat{\mathcal{U}}$ and previous $\mathcal{F}(\mathcal{W})$, achieving better smoothness in solving the iterative optimisation problem.
However, ADMM and R\_ADMM share the same convergence speed to reach the critical point.
To realise accelerated optimisation for a real-time tracker, we present the optimisation approach with the relaxed and accelerated alternating direction method of multipliers (R\_A-ADMM)~\cite{francca2018admm}, as shown in Algorithm~\ref{alg}.
Most of the steps are straight forward, except the sub-problems of optimising $\hat{\mathcal{U}}[l+1]$ and $\mathcal{W}[l+1]$.
We unfold these two sub-problems in detail.
\begin{algorithm}[t]
\begin{algorithmic}
\vspace{0.03in}
\STATE \textbf{Input:} $f\left(\hat{\mathcal{U}}\right)$, $g\left(\mathcal{W}\right)$, $h_{\rho}\left(\hat{\mathcal{U}},\mathcal{W},\mathcal{T}\right)$, penalty $\rho$, $\alpha$ and $r$
\STATE \textbf{initialise} $l=0$, $\hat{\mathcal{U}}[0]$, $\mathcal{W}[0]$,  $\mathcal{T}[0]$, $\mathcal{W}^{\prime}[0]$ and $\mathcal{T}^{\prime}[0]$
\STATE \textbf{while} convergence condition \textbf{not} satisfied, \textbf{do}
\STATE \ \ \ \ 1. $\hat{\mathcal{U}}[l+1]\leftarrow\arg\underset{\hat{\mathcal{U}}}{\min}\left\{f\left(\hat{\mathcal{U}}\right)+h_{\rho}\left(\hat{\mathcal{U}},\mathcal{W}^{\prime}[l],\mathcal{T}^{\prime}[l]\right)\right\}$
\STATE \ \ \ \ 2. $\hat{\mathcal{V}}\leftarrow\alpha\hat{\mathcal{U}}[l+1]+\left(1-\alpha\right)\mathcal{F}\left(\mathcal{W}^{\prime}[l]\right)$
\STATE \ \ \ \ 3. $\mathcal{W}[l+1]\leftarrow\arg\underset{\mathcal{W}}{\min}\left\{g\left(\mathcal{W}\right)+h_{\rho}\left(\hat{\mathcal{V}},\mathcal{W},\mathcal{T}^{\prime}[l]\right)\right\}$
\STATE \ \ \ \ 4. $\mathcal{T}[l+1]\leftarrow\mathcal{T}^{\prime}[l]+\mathcal{F}^{-1}(\hat{\mathcal{V}})-\mathcal{W}[l+1]$
\STATE \ \ \ \ 5. $\beta\leftarrow l/\left(l+r\right)$
\STATE \ \ \ \ 6. $\mathcal{T}^{\prime}[l+1]\leftarrow\mathcal{T}[l+1]+\beta(\mathcal{T}[l+1]-\mathcal{T}[l])$
\STATE \ \ \ \ 7. $\mathcal{W}^{\prime}[l+1]\leftarrow\mathcal{W}[l+1]+\beta(\mathcal{W}[l+1]-\mathcal{W}[l])$
\STATE \ \ \ \ 8. $l=l+1$
\STATE \textbf{end for}
\vspace{0.03in}
\end{algorithmic}
\caption{Optimisation via the R\_A-ADMM for problem (\ref{aug}). The relaxation parameter $\alpha\in(0,1)\cup(1,2)$ and the damping constant is $r\geq 3$.}
\label{alg}
\end{algorithm}

\noindent \textbf{Optimising} $\hat{\mathcal{U}}[l+1]$: The aim of this step is to solve the following sub-problem (we omit the iteration index $[l]$ for the sake of simplicity):
\begin{equation}
\min\left\|\sum\limits_{k=1}^{C}\hat{\mathcal{X}}^k\odot\hat{\mathcal{U}}^k-\hat{\textbf{Y}}\right\|_F^2
+\sum\limits_{k=1}^{C}\frac{\rho}{2}\left\|\hat{\mathcal{U}}^k-\hat{\mathcal{W}}^{\prime k}+\hat{\mathcal{T}}^{\prime k}\right\|^2_F.
\end{equation}
Here, $h_{\rho}$ is calculated in the frequency domain based on the Plancherel's theorem, with which a closed-form solution can be derived for each spatial unit $\hat{\mathbf{u}}_{i,j}\in\mathbb{C}^C$ as~\cite{petersen2008matrix}:
\begin{equation}\label{iter1}
\hat{\mathbf{u}}_{i,j}=\left(\mathbf{I}-\frac{\hat{\mathbf{x}}_{i,j}\hat{\textbf{x}}_{i,j}^\top}{\rho/2+\hat{\mathbf{x}}_{i,j}^\top\hat{\mathbf{x}}_{i,j}}\right)\left(\frac{\hat{\mathbf{x}}_{i,j}\hat{y}_{i,j}}{\rho}+\hat{\mathbf{w}}^{\prime}_{i,j}-\hat{\bm{\gamma}}^{\prime}_{i,j}\right),
\end{equation}
where vectors $\hat{\mathbf{w}}^{\prime}_{i,j}$ ( $\hat{\mathbf{w}}_{i,j}=\left[\hat{w}_{i,j}^1,\hat{w}_{i,j}^2,\ldots,\hat{w}_{i,j}^C\right]\in\mathbb{C}^C$), $\hat{\mathbf{x}}_{i,j}$, and $\hat{\bm{\gamma}}^{\prime}_{i,j}$ denote the $i$-th row $j$-th column units of $\hat{\mathcal{W}}^{\prime}$, $\hat{\mathcal{X}}$ and $\hat{\mathcal{T}}^{\prime}$, respectively, across all the $C$ channels.

\noindent \textbf{Optimising} $\mathcal{W}[l+1]$: The aim of this step is to solve the following sub-problem:
\begin{equation}
\min\lambda_1\sum\limits_{k=1}^{C}\left\|\mathcal{W}^k\odot\mathbf{P}\right\|_F^2
+\sum\limits_{k=1}^{C}\frac{\rho}{2}\left\|\mathcal{V}^k-\mathcal{W}^k+\mathcal{T}^{\prime k}\right\|_F^2.
\end{equation}
The above problem is channel-wise separable with an approximate solution via the mapping operator given as:
\begin{equation}
\mathcal{W}^{k}=\left(\sqrt{1+\frac{\lambda_2}{\lambda_1}}\mathbf{1}-\mathbf{P}\right)\odot\frac{\rho\mathcal{V}^k+\mathcal{T}^{\prime k}}{2\lambda_1+\rho}
\end{equation}

\subsection{Continuous Dynamical Systems}
In this section, we analyse the different approaches in the ADMM family from the perspective of equivalent continuous dynamical systems. 
Following~\cite{francca2018relax}, the equivalent form of the continuous dynamical system corresponding to our R\_A-ADMM algorithm stated in Algorithm~\ref{alg}, which is derived in 
Appendix~\ref{FirstAppendix}, is given as
\begin{equation}\label{inv}
\left(2-\alpha\right)\left[\ddot{\mathcal{W}}(t)+\frac{r}{t}\dot{\mathcal{W}}(t)\right]+\nabla\Lambda\left(\mathcal{W}(t)\right)=0,
\end{equation}
where $t=l/\sqrt{\rho}$,  $\mathcal{W}=\mathcal{W}(t)$ is the continuous limit of $\mathcal{W}[l]$, $\dot{\mathcal{W}}\equiv\frac{d\mathcal{W}}{dt}$ denotes time derivative and $\ddot{\mathcal{W}}\equiv\frac{d^2\mathcal{W}}{dt^2}$ is the acceleration. 

On the other hand, the continuous limit of algorithms R\_ADMM and ADMM withou acceleration, stated in Algorithm~\ref{alg2}, 
optimising the proposed augmented Lagrangian objective~(\ref{aug}), corresponds to the initial value problem:
\begin{equation}
\left(2-\alpha\right)\dot{\mathcal{W}}(t)+\nabla\Lambda\left(\mathcal{W}(t)\right)=0.
\end{equation}
Based on the above differential equations, the convergence rates of the above dynamical systems equivalent to R\_A-ADMM, R\_ADMM and ADMM are $\mathcal{O}\left(\left(2-\alpha\right)\left(r-1\right)^2\sigma_1^2\left(\mathbf{F}\right)/t^2\right)$, $\mathcal{O}\left(\left(2-\alpha\right)\sigma_1^2\left(\mathbf{F}\right)/t\right)$ and $\mathcal{O}\left(\sigma_1^2\left(\mathbf{F}\right)/t\right)$, respectively, where $\sigma_1\left(\mathbf{F}\right)$ is the largest singular value of $\mathbf{F}$.

Clearly, R\_A-ADMM can achieve a superior convergence rate of $\mathcal{O}(1/l^2)$ rather than the standard ADMM ($\mathcal{O}(1/l)$).
In addition, ~\cite{francca2018admm} provides a theoretical proof that the critical point of the dynamical system corresponding to R\_A-ADMM~(\ref{objre}), optimised by Algorithm~\ref{alg}, is Lyapunov stable. 

\subsection{Adaptive Initialisation via Temporal Smoothness}
To balance the characteristics of online filter update by per-frame individual learning, we propose further to accelerate the tracking process by adopting adaptive initialisation of Algorithm~\ref{alg}. 
This is accomplished by imposing temporal smoothness in order to update the optimised filters from the last learning stage to the current one.
Different from existing techniques~\cite{Danelljan2016Beyond} that establish specific temporal appearance models, we achieve an efficient temporal connection by a simple transfer learning without any additional complexity.
It is intuitive that an optimised model for one selected frame retains its validity around its neighbouring frames. We therefore start the filter learning process for frame $k$ from the filter value at frame $k-1$.
It should be noted that such adaptive initialisation enables not only acceleration, but also increases robustness of the optimisation process.

\section{Evaluation}\label{experiment}
In the section, we perform both qualitative and quantitative experiments to evaluate the tracking performance of the proposed ACFT method. The algorithm implementation is presented first, detailing the parameters setting and hardware particulars, followed by the tracking datasets and the corresponding evaluation metrics.
We then report a component analysis of the proposed accelerated method, illustrating the impact of different iterative approaches and deep CNN representation configurations.
The tracking performance is also compared with the state-of-the-art methods in terms of the overall, as well as attribute metrics.

\subsection{Implementation Details}\label{detail}
We implement the proposed tracking algorithm in Matlab 2018a on a platform with Intel(R) Xeon(R) E5-2637 CPU and GTX TITAN X GPU. 
We use the VLFeat toolbox for HOG features, lookup table from the original paper~\cite{Weijer2009Learning} for CN representation, and MatConvNet~\footnote{http://www.vlfeat.org/matconvnet/} for deep CNN features.
We configure the HOG descriptor as a 9-orientation operator with a cell size $4\times 4$, and the CN feature with a cell size $4\times 4$, to unify the hand-crafted feature stride as 4.
The related parameters of the proposed ACFT are set as follows:
the area ratio between search window and target region, $\beta=16$, regularisation parameters, $\lambda_1=10$, $\lambda_2=100$,
the penalty, $\rho=1.0$, 
the relaxation  parameter, $\alpha=1.10$,
the damping constraint, $r=4$.
The maximal iteration number is $8$ with the convergence condition as $\left|\Lambda\left(\mathcal{W}[l]\right)-\Lambda\left(\mathcal{W}[l-1]\right)\right|/\left(N\times N\times C\right)<5\times10^{-7}$.
All the parameters are fixed for all the datasets.
The raw results and MATLAB code will be publicly available at GitHub (\url{https://github.com/XU-TIANYANG}).

\begin{figure*}[!t]
\begin{center}
   \includegraphics[trim={30mm 0mm 30mm 0mm},clip,width=.48\linewidth]{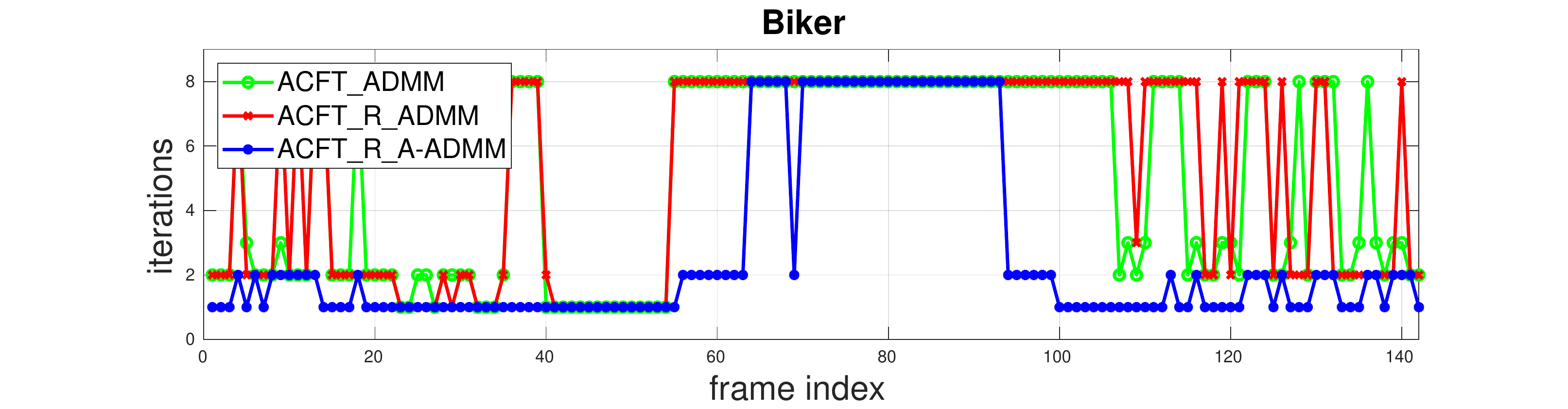}
   \includegraphics[trim={30mm 0mm 30mm 0mm},clip,width=.48\linewidth]{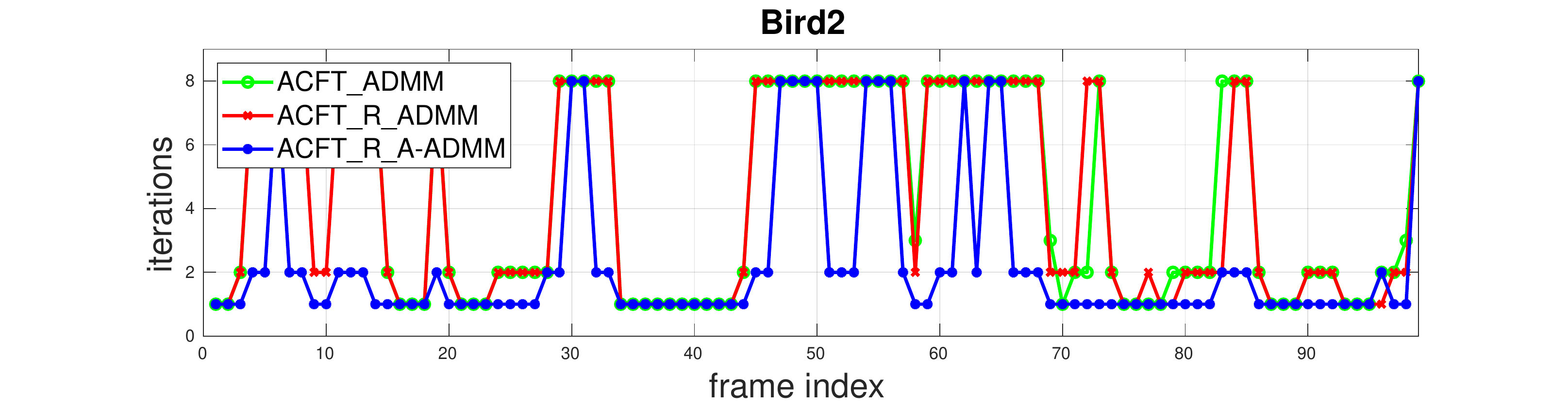}\\
      \includegraphics[trim={30mm 0mm 30mm 0mm},clip,width=.48\linewidth]{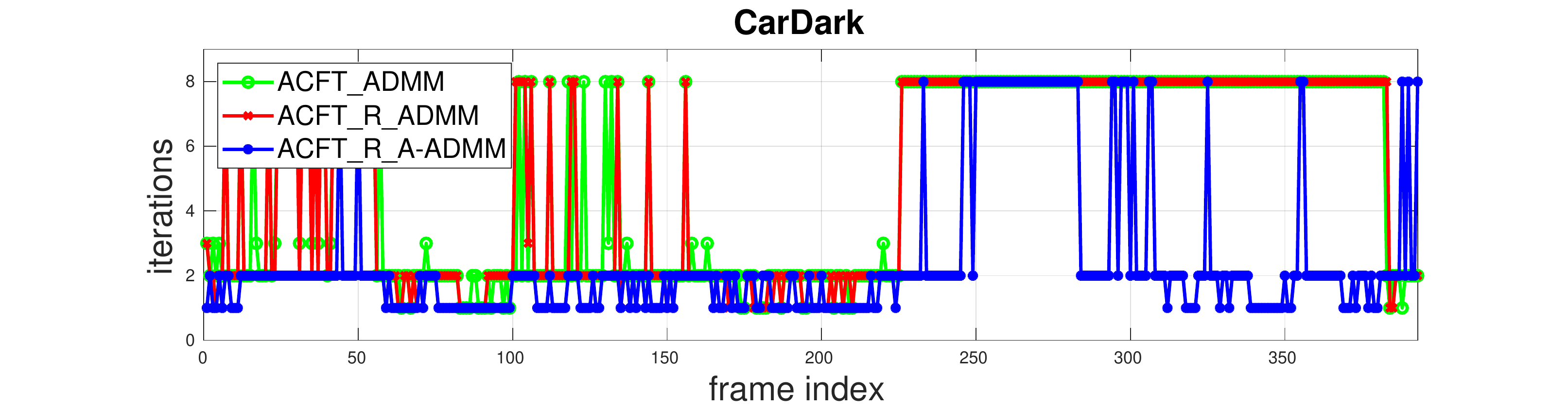}
   \includegraphics[trim={30mm 0mm 30mm 0mm},clip,width=.48\linewidth]{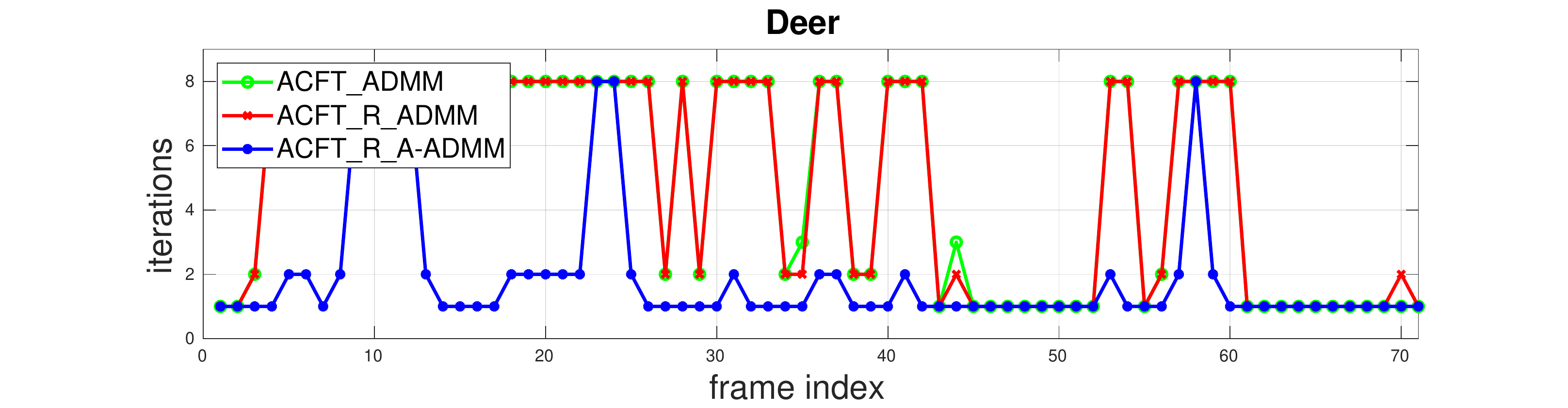}\\
      \includegraphics[trim={30mm 0mm 30mm 0mm},clip,width=.48\linewidth]{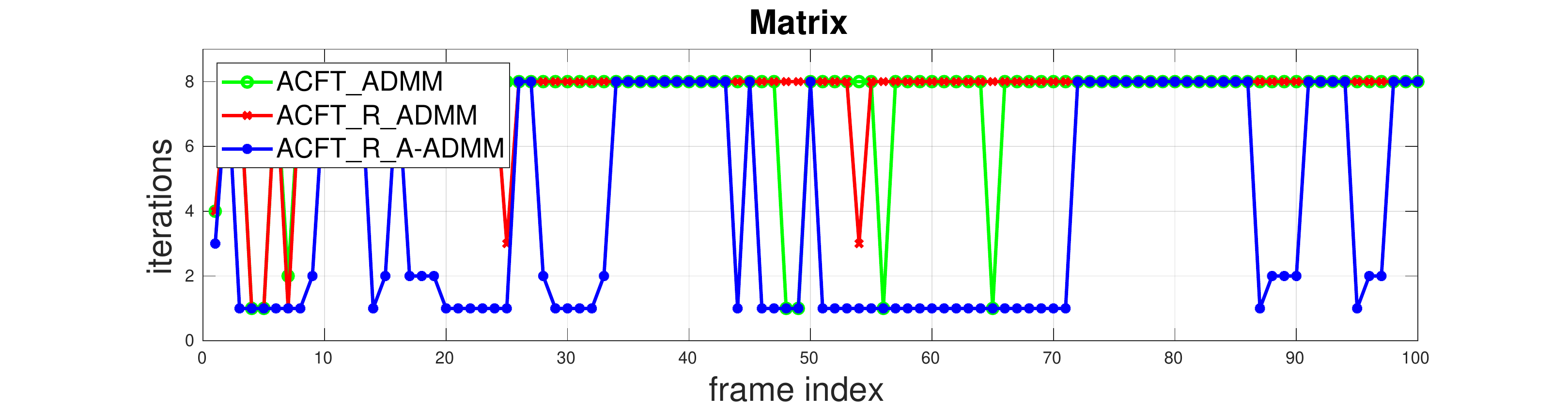}
   \includegraphics[trim={30mm 0mm 30mm 0mm},clip,width=.48\linewidth]{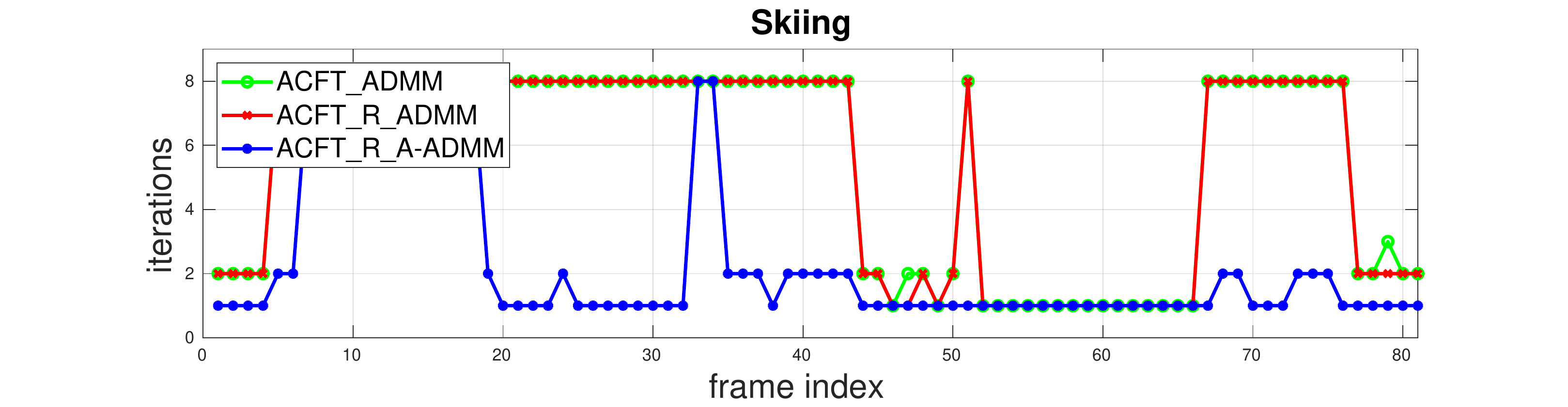}\\
\end{center}
   \caption{The number of iterations required for $\mathcal{W}$ to converge using R\_A-ADMM, R\_ADMM and ADMM for ACFT on sequences~\cite{Wu2015Object} \textit{Biker},  \textit{Bird2}, \textit{CarDark}, \textit{Deer}, \textit{Matrix}, and \textit{Skiing}, respectively.}
   \label{Impact_of_L}
\end{figure*}

\begin{table}[!b]
\footnotesize
\renewcommand{\arraystretch}{1.1}
\caption{The tracking results of ACFT obtained on OTB2015 by different optimisation approaches using CNN representations.}
\label{cnn}
\centering
\begin{tabular}{cl|ccc|c}
\hline
&& \textbf{DP} & \textbf{OP} & \textbf{AUC} & \textbf{FPS}\\
\hline
\hline
\multirow{ 3}{*}{ADMM}&AlexNet & 90.40 & 86.01 & 68.30 & 15.1\\
&VGG-16 & 90.13 & 85.96 & 68.10 & 2.7\\
&ResNet-50 & 91.08 & 86.88 & 69.21 & 3.1 \\
\hline
\multirow{ 3}{*}{R\_ADMM}&AlexNet & 90.52 & 85.91 & 68.49 & 16.3\\
&VGG-16 & 90.27 & 86.23 & 68.51 & 2.9\\
&ResNet-50 & 91.03 & 87.19 & 69.43 & 3.5 \\
\hline
\multirow{ 3}{*}{R\_A-ADMM}&AlexNet & 90.70 & 86.33 & 68.80 & \textbf{18.8}\\
&VGG-16 & 90.63 & 86.54 & 68.53 & 3.2\\
&ResNet-50 & \textbf{91.22} & \textbf{87.22} & \textbf{69.52} & 4.1 \\
\hline 
\end{tabular}
\end{table}
\subsection{Datasets and Evaluation Metrics}
We perform the evaluation on several standard datsets,~\textit{i.e.}, OTB2013~\cite{Wu2013Online}, TB50~\cite{Wu2015Object}, OTB2015~\cite{Wu2015Object} and VOT2017~\cite{Kristan2017a}.
OTB2013, TB50 and OTB2015 are widely used tracking datasets that respectively contain 51, 50 and 100 annotated challenging video sequences with 11 sequence attributes. 
As our ACFT belongs to a deterministic algorithm, One Pass Evaluation (\textbf{OPE}) protocol~\cite{Wu2013Online} is applied to evaluate the performance of different trackers on the above three datasets. 
The precision and success plots report the tracking results based on centre location error and bounding box overlap. 
In addition, the Area Under Curve (\textbf{AUC}), Centre Location Error (\textbf{CLE}), Overlap Precision (\textbf{OP}, percentage of overlap ratios exceeding 0.5) and Distance Precision (\textbf{DP}, percentage of location errors within 20 pixels) provide objective numerical values for performance comparison.
The speed of a tracking algorithm is measured in Frames Per Second (\textbf{FPS}).
We compare our ACFT with a number of state-of-the-art DCF trackers, including MetaT~\cite{park2018meta} (ECCV18), MCPF~\cite{zhang2017multi} (CVPR17), CREST~\cite{song-iccv17-CREST} (ICCV17), BACF~\cite{Galoogahi2017Learning} (ICCV17), CFNet~\cite{valmadre2017end} (CVPR17), STAPLE\_CA~\cite{mueller2017context} (CVPR17), ACFN~\cite{choi2017attentional} (CVPR17), CSRDCF~\cite{Lukezic2017Discriminative} (CVPR17), C-COT~\cite{Danelljan2016Beyond} (ECCV16), Staple~\cite{Bertinetto2016Staple} (CVPR16), SRDCF~\cite{Danelljan2015Learning} (ICCV15), KCF~\cite{Henriques2015High} (TPAMI15), SAMF~\cite{li2014scale} (ECCVW14) and DSST~\cite{danelljan2017discriminative} (TPAMI17).

The VOT2017 benchmark consists of 60 challenging video sequences. 
We employ the expected average overlap (\textbf{EAO}), \textbf{Accuracy} and \textbf{Robustness} to evaluate the baseline and real-time performance.
We compare our ACFT with the trackers of top performance in~\cite{Kristan2017a}, \textit{i.e.}, LSART (CVPR2018), CFWCR (ICCV17), CFCF (TPAMI18), ECO (CVPR17), CSRDCF++ (CVPR17), ECO$\_$HC (CVPR17), SiamFC (ECCV16).

\subsection{Component Analysis}
First, we obtain the tracking results of our ACFT using different optimisation methods as well as different deep CNN features.
Standard ADMM, R\_ADMM and R\_A-ADMM are compared in terms of the tracking accuracy and iteration numbers.
We also explore the impact of different deep CNN features, namely AlexNet, VGG-16 and ResNet-50, on the tracking performance for each optimisation approach.

The different optimisation methods are compared in terms of convergence speed in Fig.~\ref{Impact_of_L}.
We record the number of iterations required for the estimate (filters) to converge in each frame during the tracking for the sequences~\cite{Wu2015Object} \textit{Biker}, \textit{Bird2}, \textit{CarDark}, \textit{Deer}, \textit{Matrix}, and \textit{Skiing}.
R\_A-ADMM outperforms ADMM and R\_ADMM in terms of the required number of iterations to reach converge.
It should be noted that a maximal number of iterations (8 in our experiment) is pre-defined to balance the learning speed and objective convergence. The results are consistent with the findings of the theoretical analysis of their corresponding continuous dynamical systems~\cite{francca2018admm}.

\begin{figure*}[!t]
\centering
\includegraphics[width=0.32\linewidth]{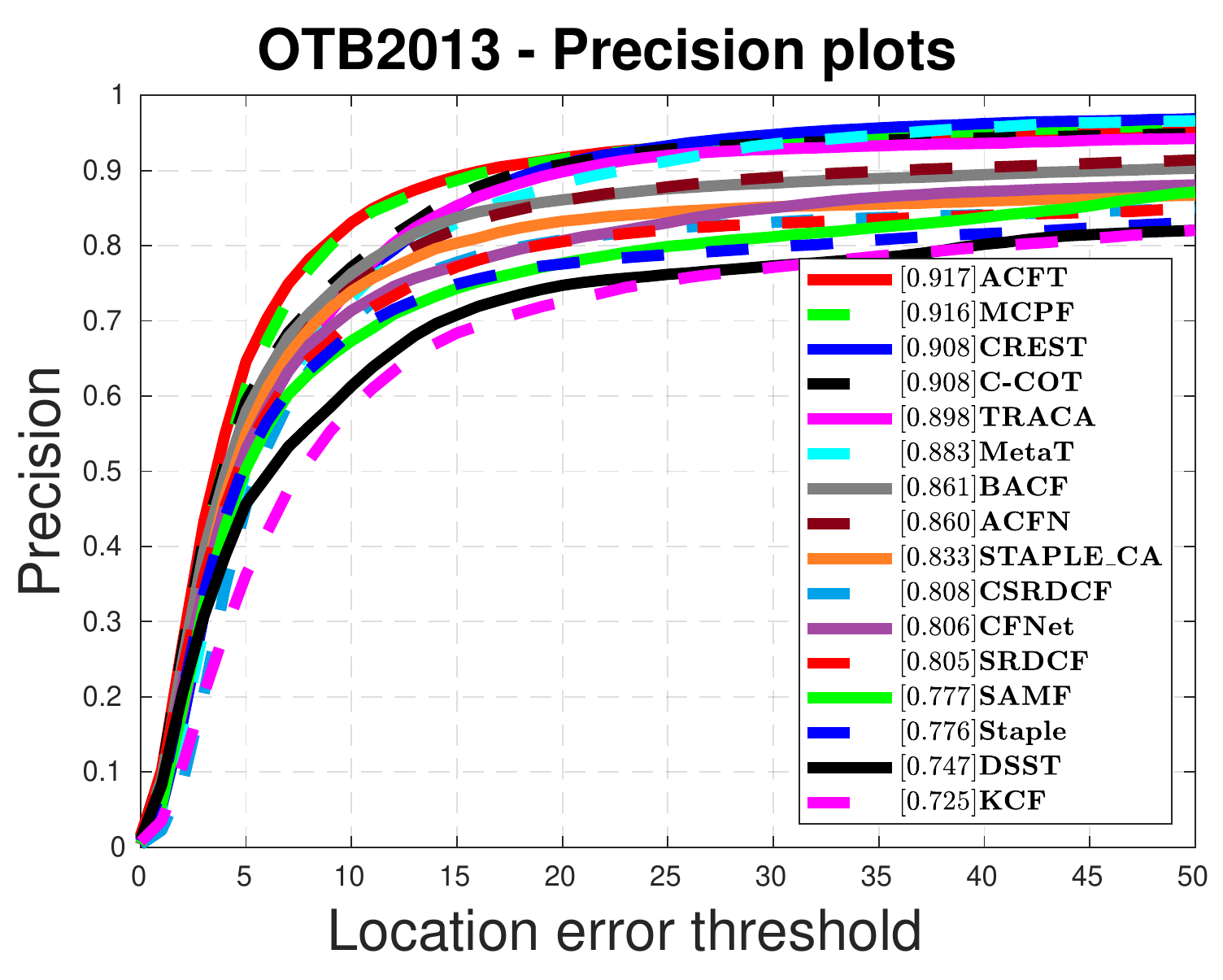}
\includegraphics[width=0.32\linewidth]{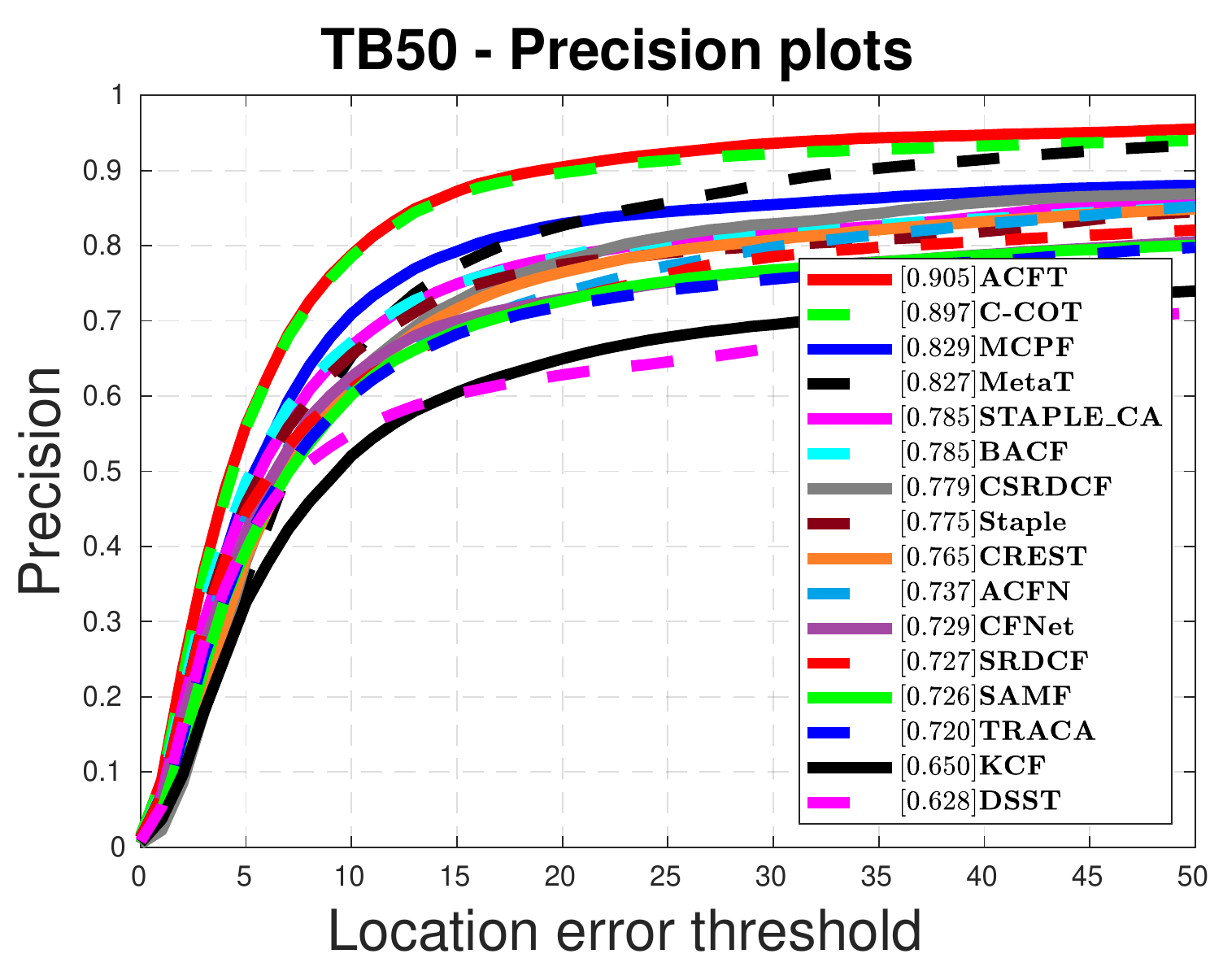}
\includegraphics[width=0.32\linewidth]{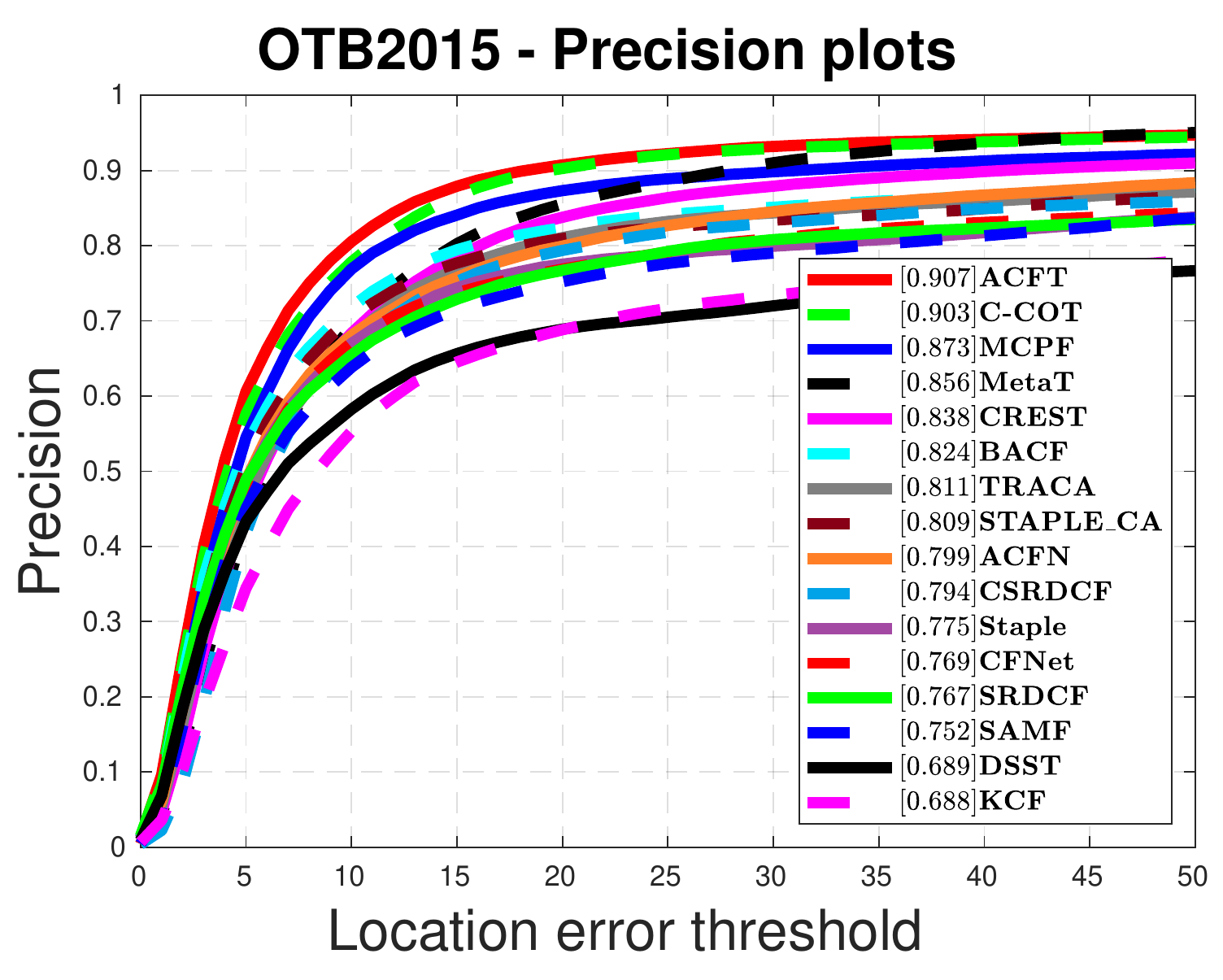}
\\
\includegraphics[width=0.32\linewidth]{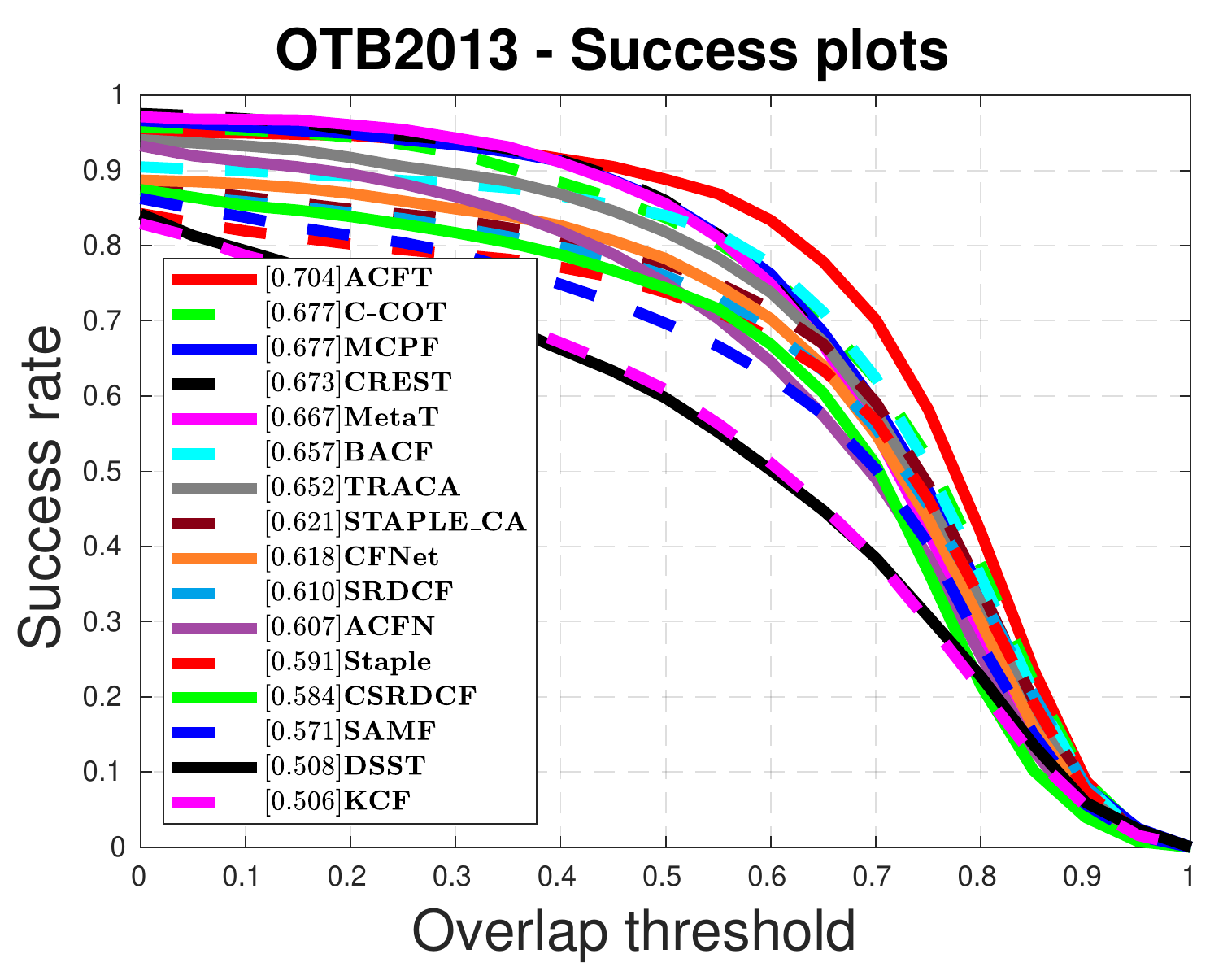}
\includegraphics[width=0.32\linewidth]{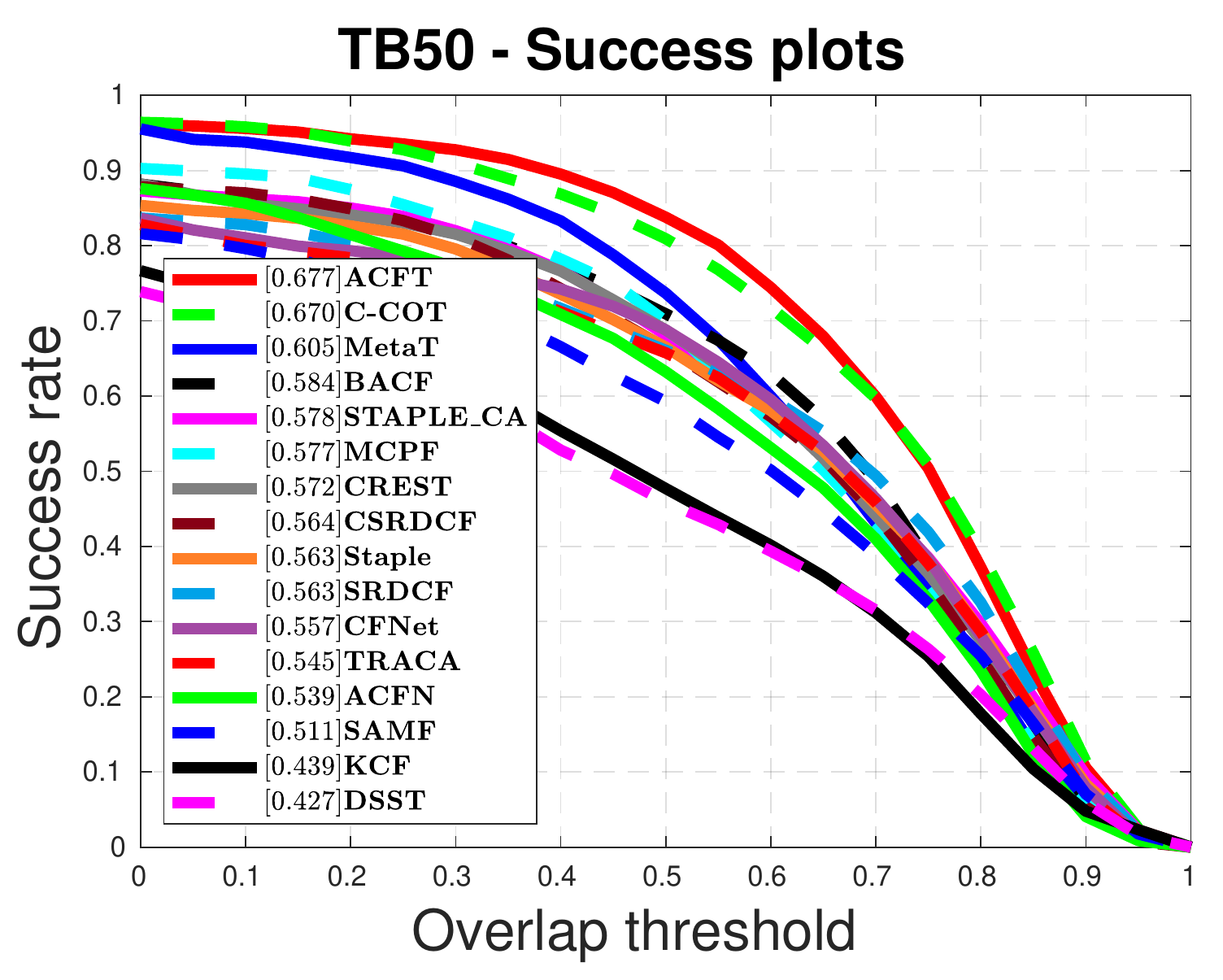}
\includegraphics[width=0.32\linewidth]{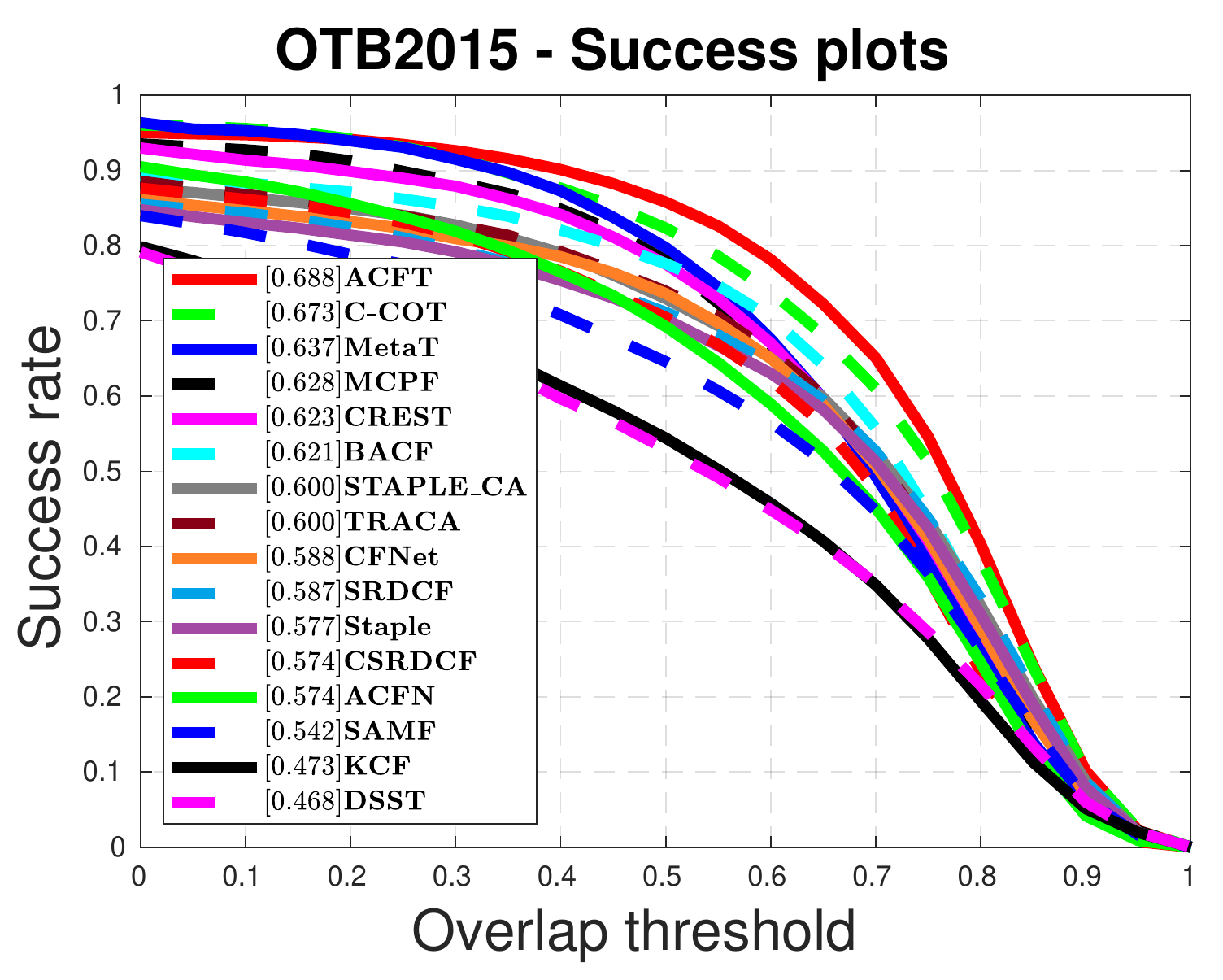}
\caption{The results of evaluation on OTB2013, TB50 and OTB2015. The precision plots (\textit{first row}) with \textbf{DP} in the legend and the success plots (\textit{second row}) with \textbf{AUC} in the legend are presented.}
\label{otbfig}
\end{figure*}

\begin{table*}[!t]
\footnotesize
\renewcommand{\arraystretch}{1.3}
\caption{A comparison of our ACFT with state-of-the-art methods on OTB2013, TB50 and OTB2015 in terms of \textbf{OP} and \textbf{CLE}. The top three results are highlighted in {\color{red}{red}}, {\color{blue}{blue}} and {\color{brown}{brown}} fonts.}
\label{otb}
\centering
\setlength{\leftskip}{-100pt}
\begin{tabular}{c|l|cccccccc}
\hline
 & & KCF & SAMF & DSST & SRDCF  & Staple  & CSRDCF & ACFN&\\
\hline
\hline
\multirow{ 3}{*}{$\begin{matrix}\textrm{Mean\  \textbf{OP/CLE}}\\($\%$/\textrm{pixels})\end{matrix}$} &\textbf{OTB2013}~\cite{Wu2013Online} &  60.8/36.3 & 69.6/29.0 & 59.7/39.2 & 76.0/36.8  & 73.8/31.4 &  74.4/31.9 & 75.0/18.7&\\
&\textbf{TB50}~\cite{Wu2015Object}    & 47.7/54.3 & 59.3/40.5 & 45.9/59.5 & 66.1/42.7  & 66.5/32.3 &  66.4/30.3 & 63.2/32.1&\\
&\textbf{OTB2015}~\cite{Wu2015Object}    & 54.4/45.1 & 64.6/34.6 & 53.0/49.1 & 71.1/39.7  & 70.2/31.8 &  70.5/31.1 & 69.2/25.3&\\
 Mean \textbf{FPS} & & 82.7 & 11.5 & 25.6 & 2.7  & 5.2 & 4.6 & 13.8&\\
\hline
& & STAPLE\_CA & CFNet & BACF & CREST & MCPF & MetaT & C-COT & \textbf{ACFT}\\
\hline
\hline
\multirow{ 3}{*}{$\begin{matrix}\textrm{Mean\  \textbf{OP/CLE}}\\($\%$/\textrm{pixels})\end{matrix}$}  &\textbf{OTB2013} & 77.6/29.8 & 78.3/35.2 & 84.0/26.2 & {\color{blue}{\textbf{86.0}}}/{\color{red}{\textbf{10.2}}} & {\color{brown}{\textbf{85.8}}}/{\color{blue}{\textbf{11.2}}} &  85.6/{\color{brown}{\textbf{11.5}}} & 83.7/15.6 & {\color{red}{\textbf{88.8}}}/17.4 \\
&\textbf{TB50}    & 68.1/36.3 & 68.8/36.7 & 70.9/30.3 & 68.8/32.6 & 69.9/30.9 & {\color{brown}{\textbf{73.7}}}/{\color{brown}{\textbf{17.0}}} & {\color{blue}{\textbf{80.9}}}/{\color{blue}{\textbf{12.3}}} & {\color{red}{\textbf{83.8}}}/{\color{red}{\textbf{12.2}}}\\
&\textbf{OTB2015}   & 73.0/33.1 & 73.6/36.0 & 77.6/28.2 & 77.6/21.2 & 78.0/20.9  & {\color{brown}{\textbf{79.8}}}/{\color{blue}{\textbf{14.2}}} & {\color{blue}{\textbf{82.3}}}/{\color{red}{\textbf{14.0}}} & {\color{red}{\textbf{86.3}}}/{\color{brown}{\textbf{14.9}}}\\
Mean \textbf{FPS} & & 18.2 & 8.7 & 20.3 & 10.1 & 0.5  & 0.8 & 1.3 & 18.8\\
\hline
\end{tabular}
\end{table*}

\begin{table*}[!t]
\footnotesize
\renewcommand{\arraystretch}{1.3}
\caption{Tracking results on VOT2017. The top three results are highlighted in {\color{red}{red}}, {\color{blue}{blue}} and {\color{brown}{brown}} fonts.}
\label{vot}
\centering
\setlength{\leftskip}{-50pt}
\begin{tabular}{l|l|cccccccccc|c}
\hline
& & ECO\_HC & SiamFC & CSRDCF++ & ECO & CFCF & CFWCR & LSART & \textbf{ACFT}\\
\hline
\hline
\multirow{ 3}{*}{Baseline}&\textbf{EAO} & 0.238 & 0.188 & 0.229 & 0.280 & 0.286 & {\color{brown}{\textbf{0.303}}} & {\color{red}{\textbf{0.323}}} & {\color{blue}{\textbf{0.317}}}\\
&\textbf{Accuracy} & 0.494 & {\color{brown}{\textbf{0.502}}} & 0.453 & 0.483 & {\color{blue}{\textbf{0.509}}} & 0.484 & 0.493 & {\color{red}{\textbf{0.522}}}\\
&\textbf{Robustness} & 0.435 & 0.585 & 0.370 & 0.276 & 0.281 & {\color{brown}{\textbf{0.267}}} & {\color{red}{\textbf{0.218}}} & {\color{blue}{\textbf{0.238}}}\\
\hline 
\multirow{ 3}{*}{Real-time}&\textbf{EAO} & 0.177 & {\color{brown}{\textbf{0.182}}} & {\color{blue}{\textbf{0.212}}} & 0.078 & 0.059 & 0.062 & 0.055 & {\color{red}{\textbf{0.267}}}\\
&\textbf{Accuracy} & {\color{brown}{\textbf{0.494}}} & {\color{blue}{\textbf{0.502}}} & 0.459 & 0.449 & 0.339 & 0.393 & 0.386 & {\color{red}{\textbf{0.517}}}\\
&\textbf{Robustness} & {\color{brown}{\textbf{0.571}}} & 0.604 & {\color{blue}{\textbf{0.398}}} & 1.466 & 1.723 & 1.864 & 1.971 & {\color{red}{\textbf{0.291}}}\\
\hline 
Unsupervised&\textbf{AO} & 0.335 & 0.345 &0.298 & {\color{brown}{\textbf{0.402}}} & 0.380 & 0.370 & {\color{blue}{\textbf{0.437}}} & {\color{red}{\textbf{0.449}}}\\
\hline 
\end{tabular}
\end{table*}

\begin{figure*}[!t]
   \includegraphics[width=0.245\linewidth]{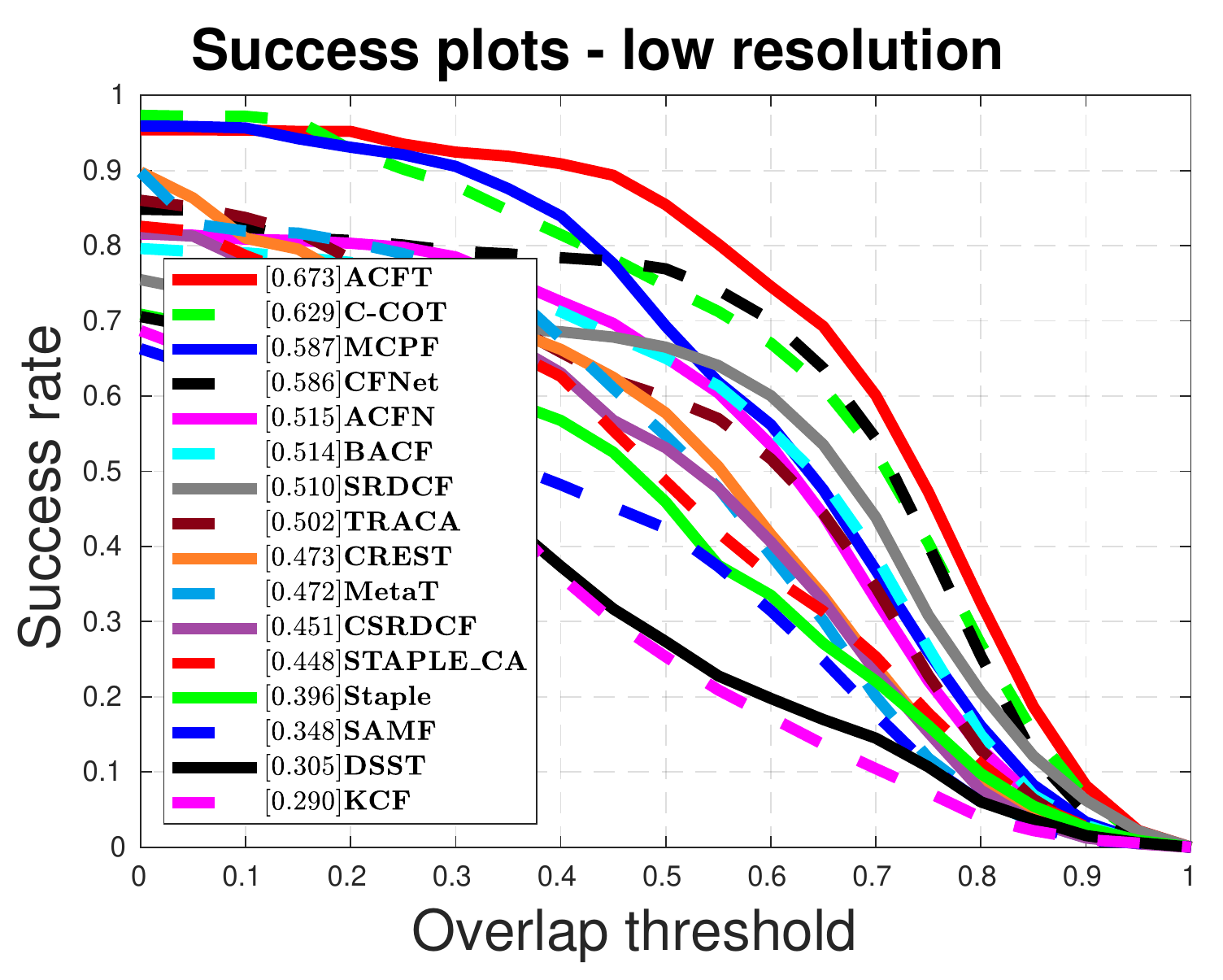}
   \includegraphics[width=0.245\linewidth]{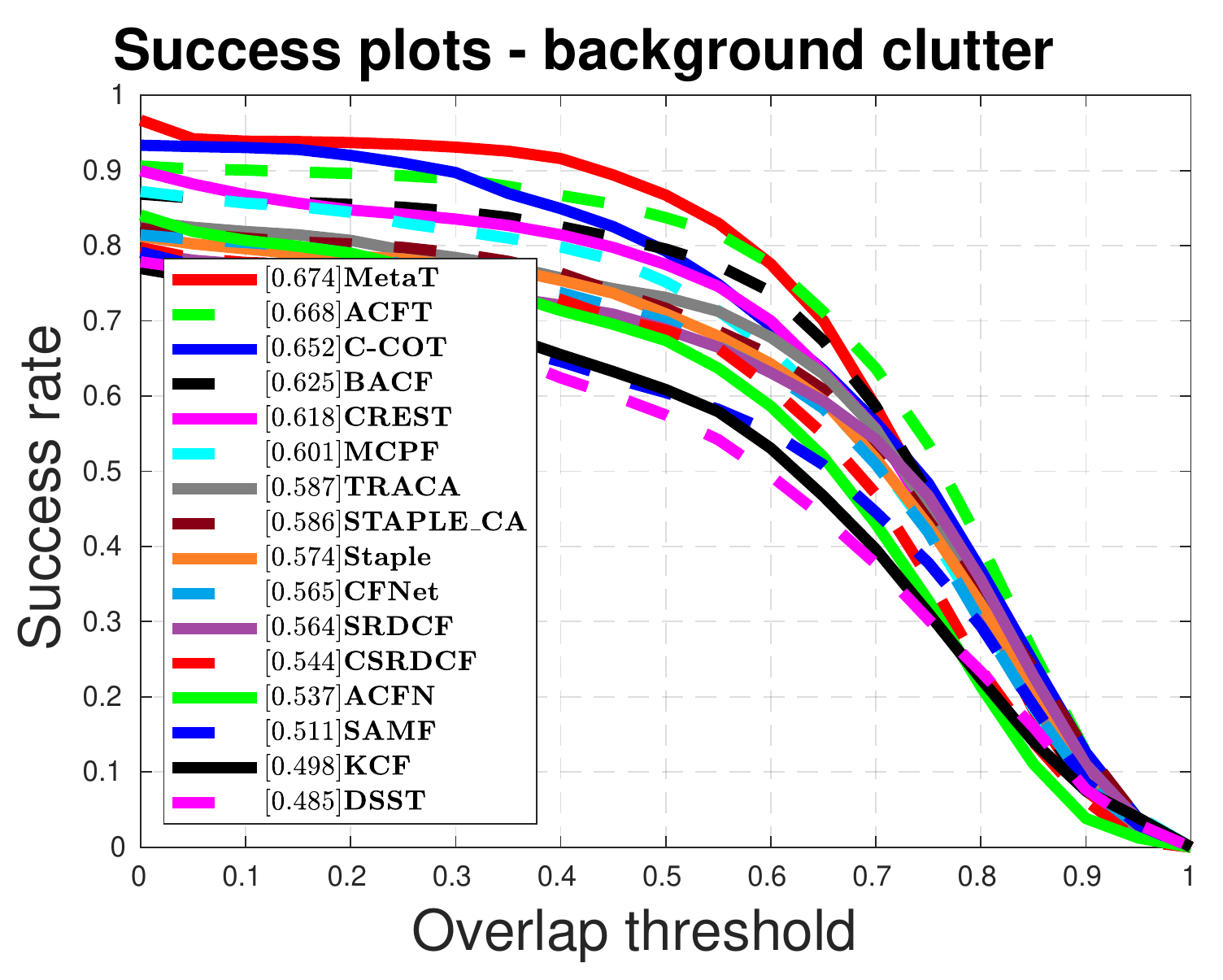}
   \includegraphics[width=0.245\linewidth]{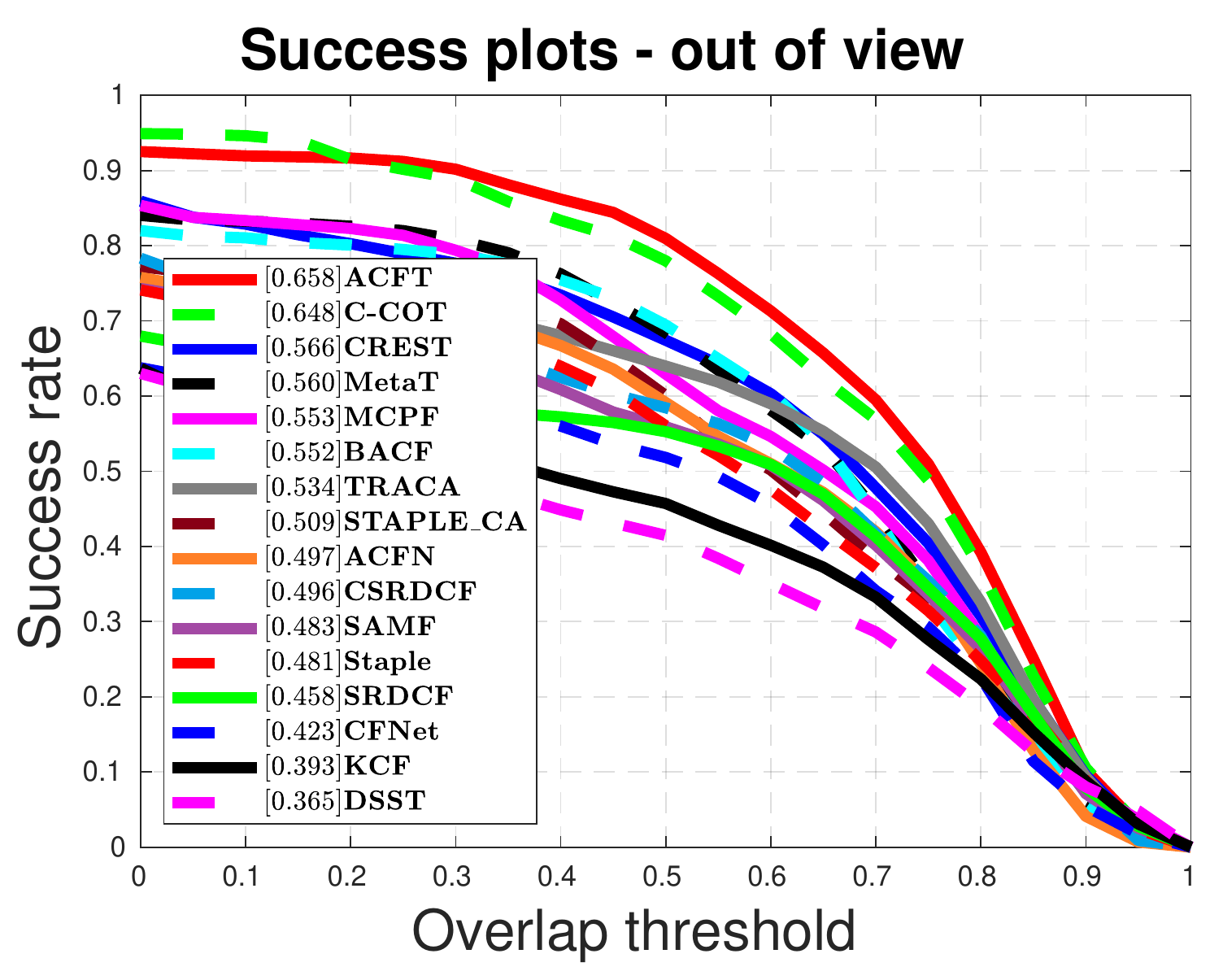}
   \includegraphics[width=0.245\linewidth]{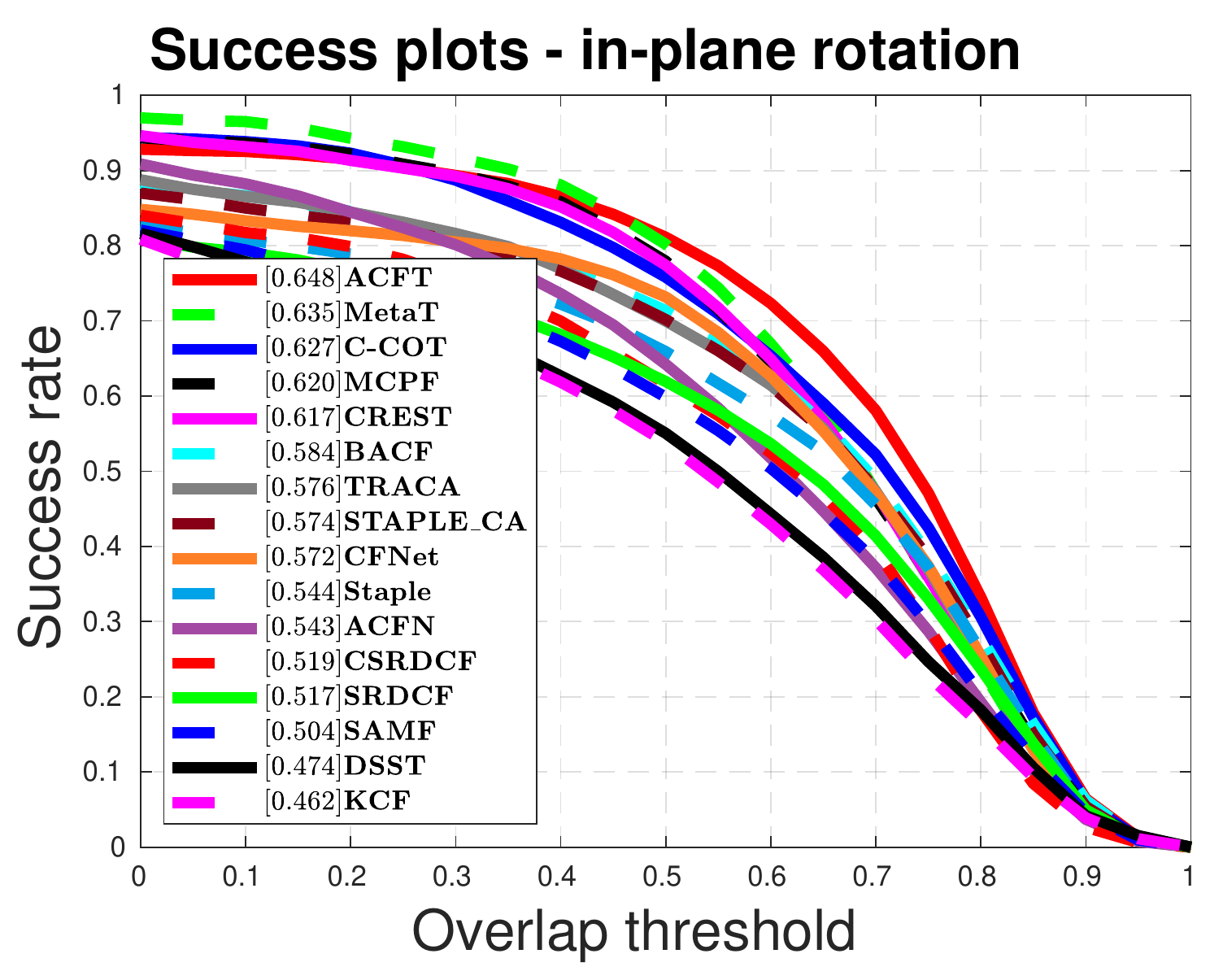}\\
   \includegraphics[width=0.245\linewidth]{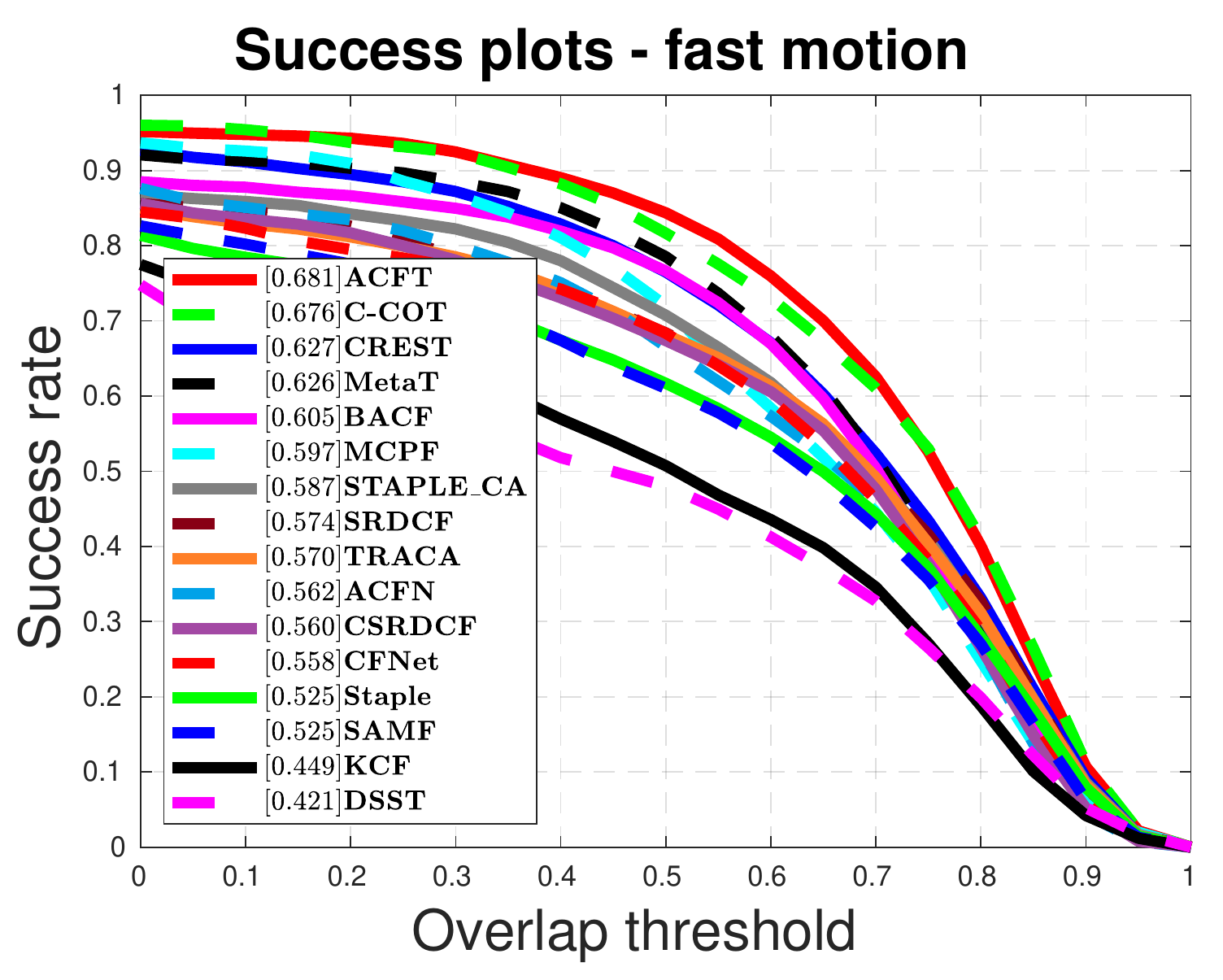}
   \includegraphics[width=0.245\linewidth]{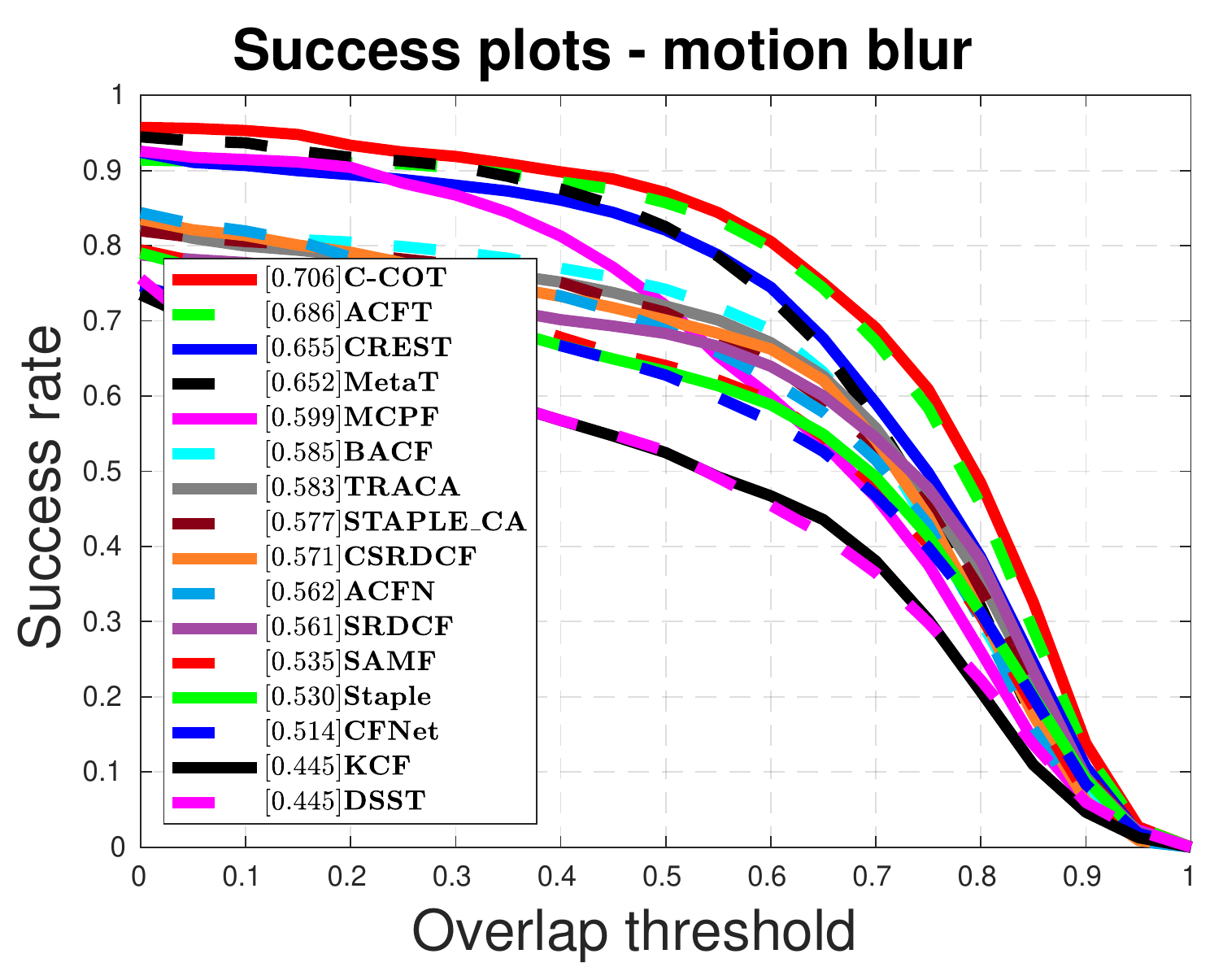}
   \includegraphics[width=0.245\linewidth]{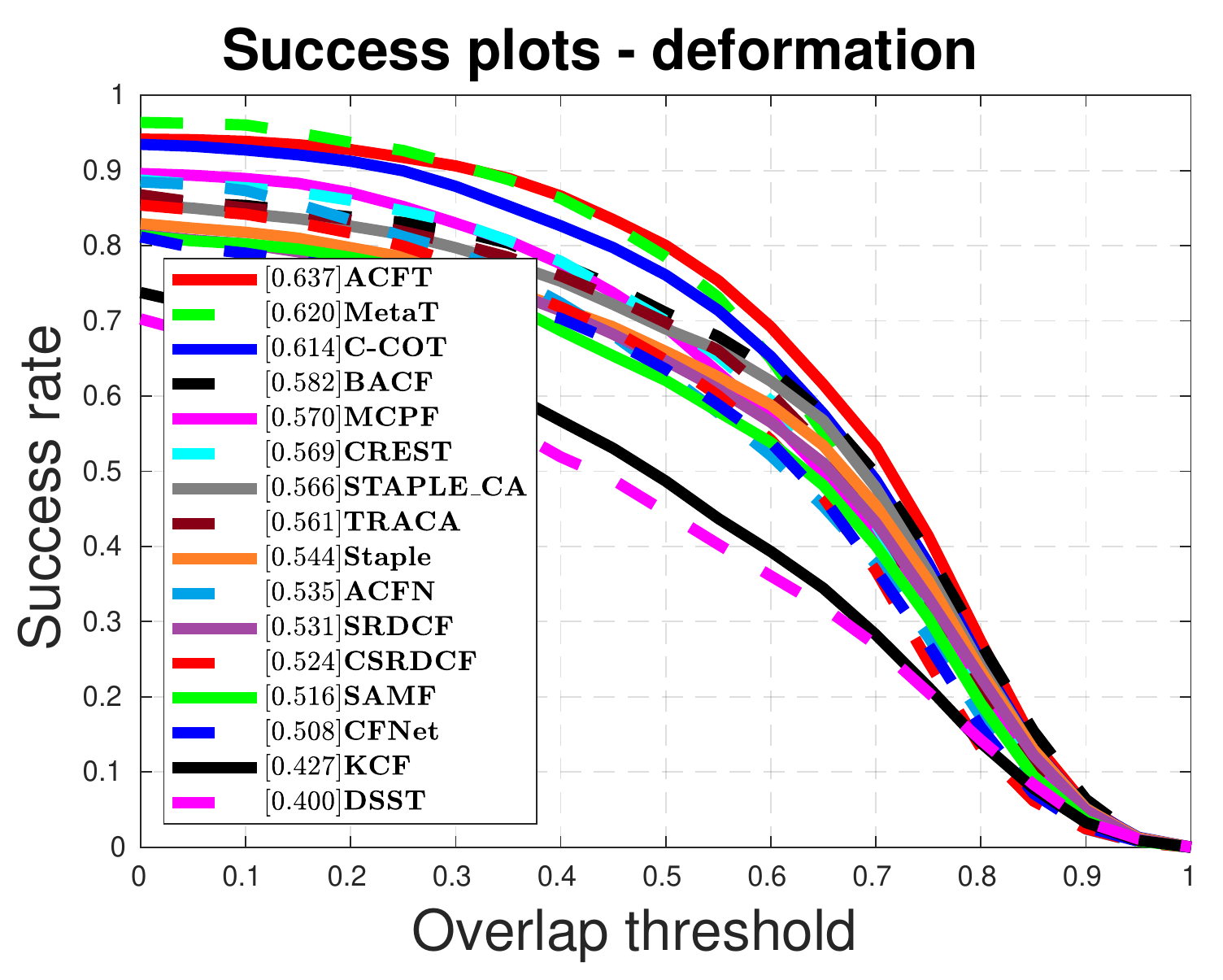}
   \includegraphics[width=0.245\linewidth]{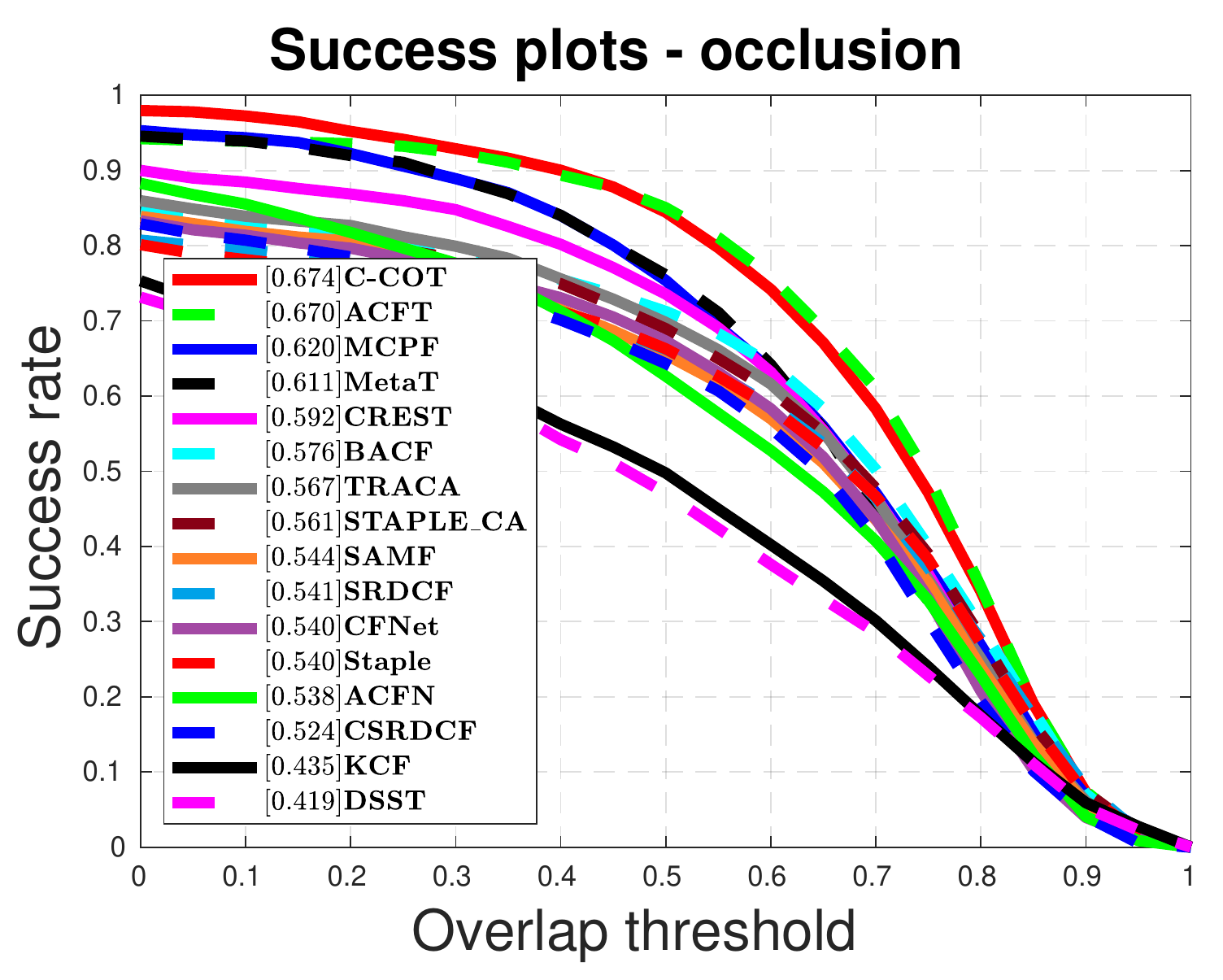}\\
    \includegraphics[width=0.245\linewidth]{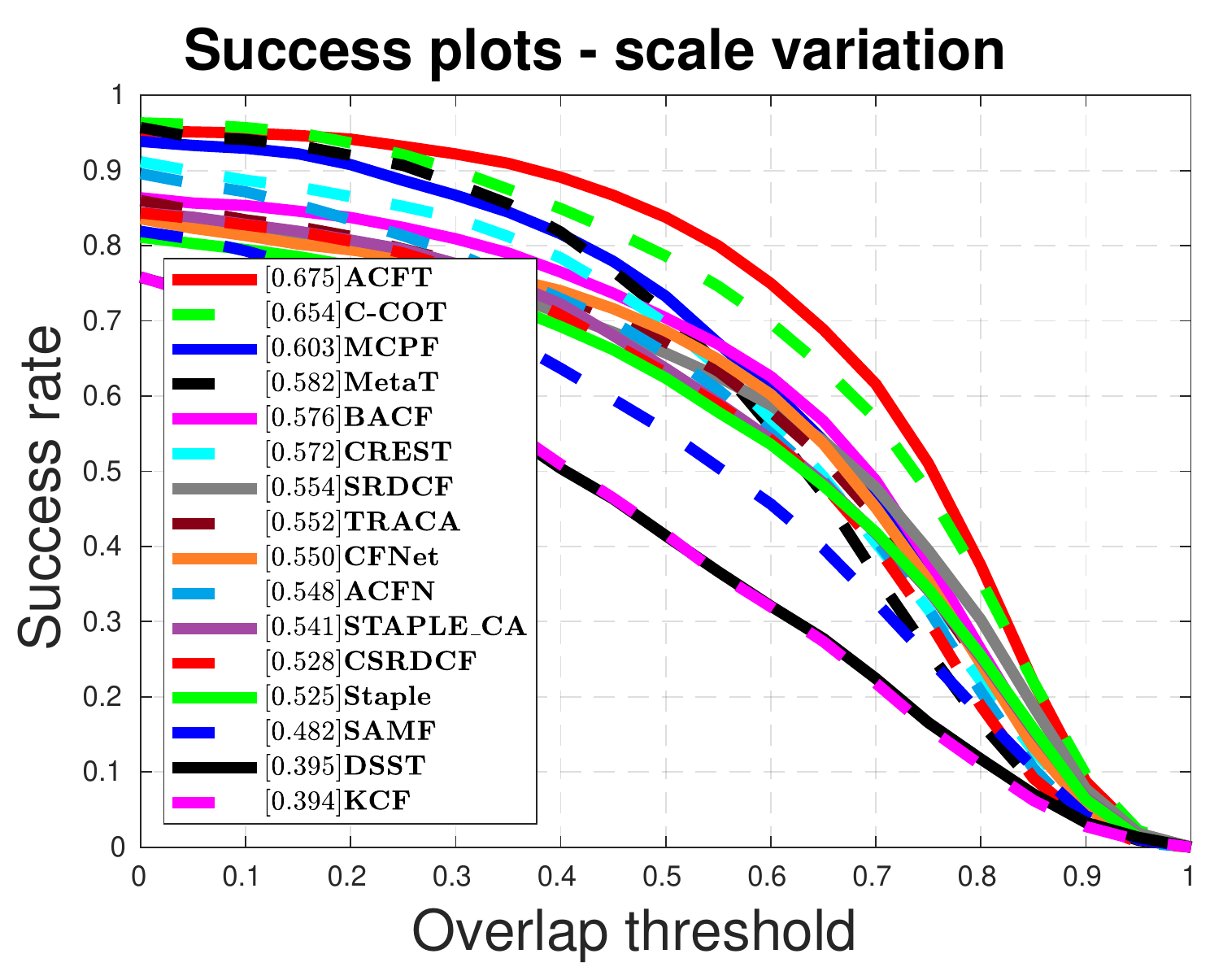}
   \includegraphics[width=0.245\linewidth]{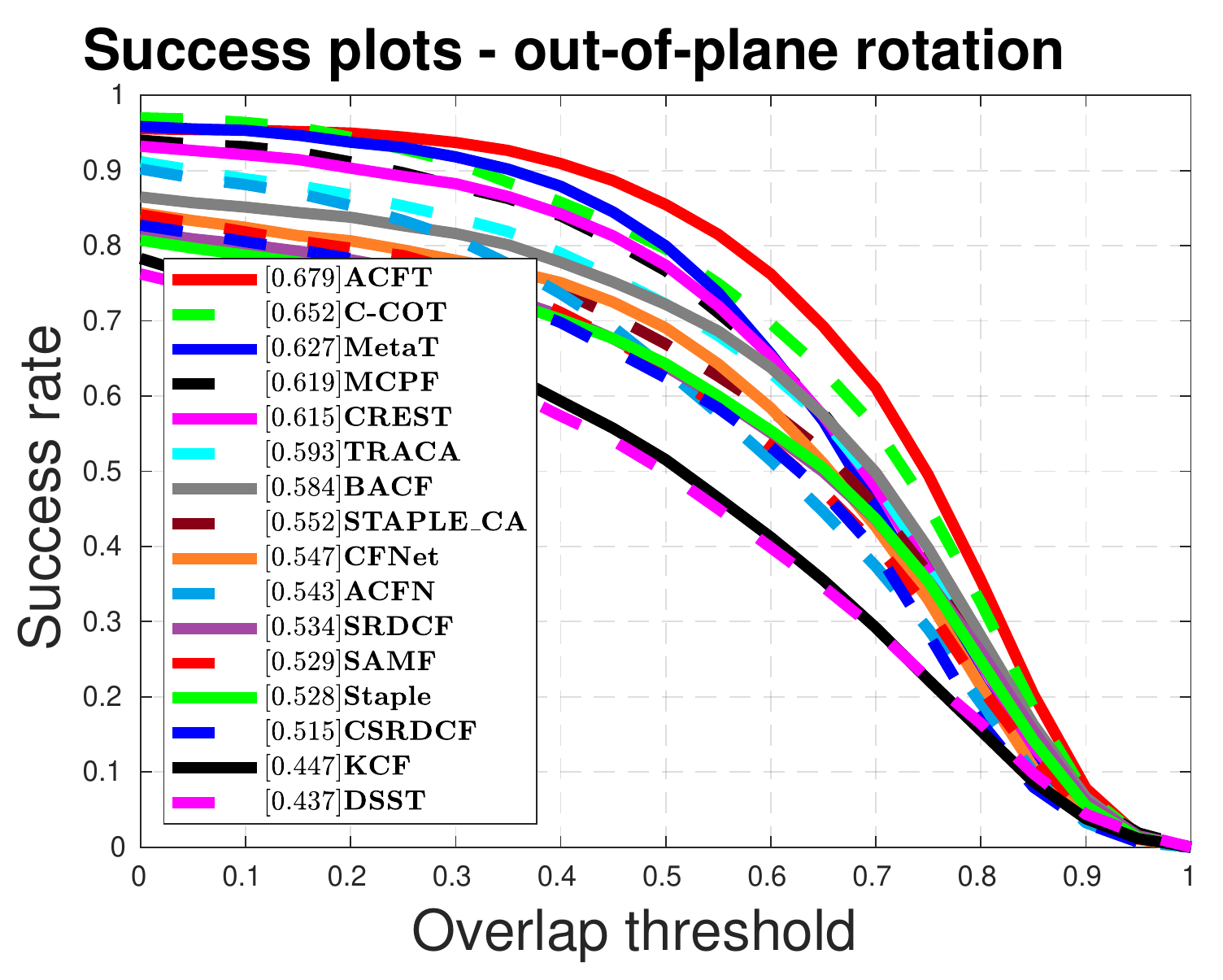}
   \includegraphics[width=0.245\linewidth]{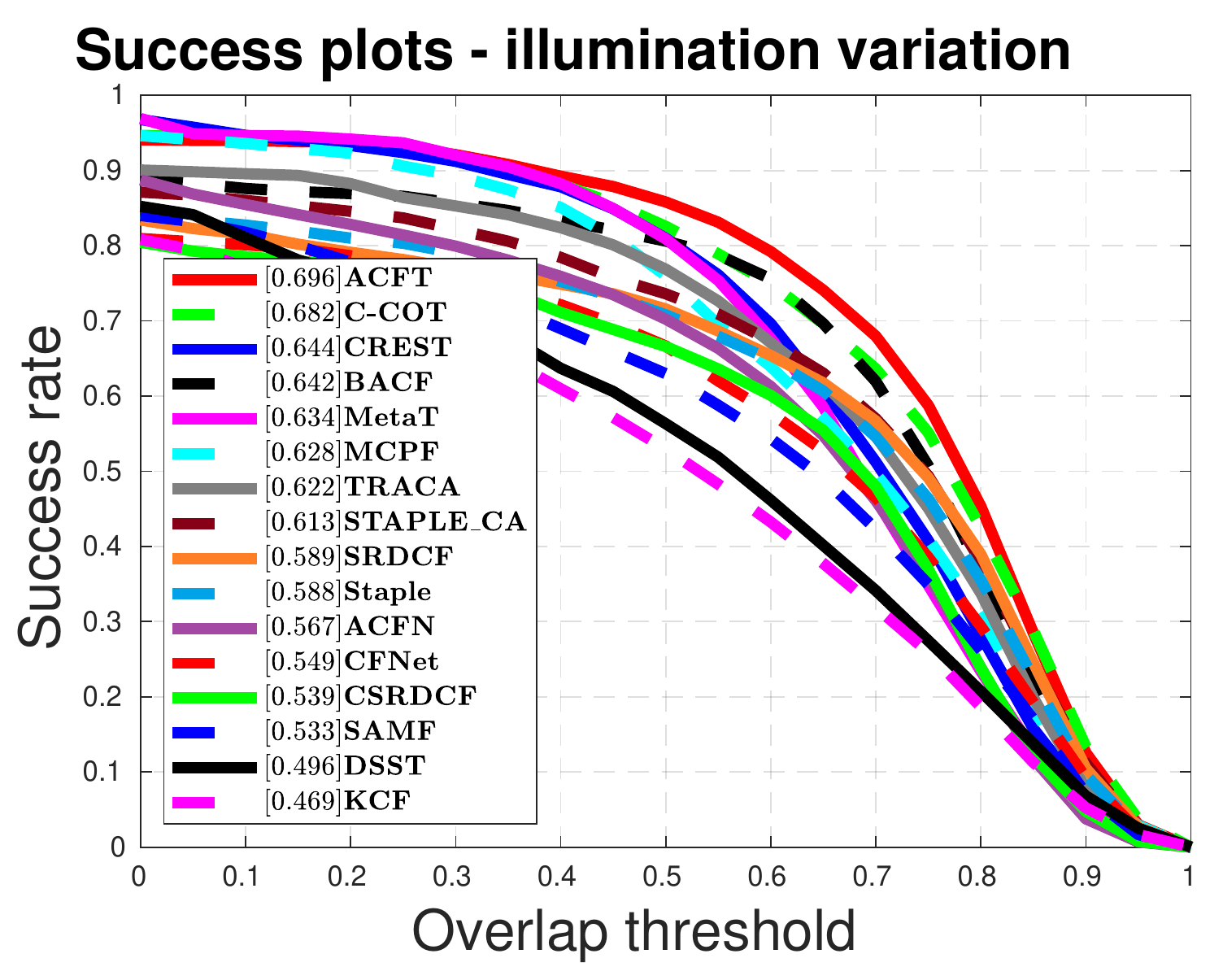}
   \caption{Success plots based on the tracking results obtained on OTB2015 in 11 sequence attributes,~\textit{i.e.}, low resolution, background clutter, out of view, in-plane rotation, fast motion, motion blur, deformation, occlusion, scale variation, out-of-plane rotation and illumination variation.}\label{attributes}
\end{figure*}

The tracking performance obtained with different deep CNN features and optimisation algorithms is reported in Table~\ref{cnn}.
For deep CNN features, we use the feature map output from the convolutional layers 'x30' and 'res4ex' as deep feature representations from VGG-16 and ResNet-50, respectively.
The configuration of hand-crafted features is fixed as detailed in Subsection~\ref{detail}.
From Table~\ref{cnn}, first, it is clear that the tracking performance can be improved by using more powerful deep CNN features for all the used optimisation approaches, while sacrificing the speed.
The AlexNet-based ACFT\_R\_A-ADMM outperforms ACFT\_R-ADMM and ACFT\_ADMM in terms of \textbf{DP}, \textbf{OP} and \textbf{AUC}.
The gap between the optimal \textbf{OP} and \textbf{DP} values obtained by ADMM, R\_ADMM and R\_A-ADMM with the same CNN features is less than $1\%$, demonstrating that the whole family of ADMM algorithms achieve effective optimisation of the objective function in Equ.~\ref{aug}.
Specifically, the ACFT\_R\_A-ADMM configuration with ResNet-50 achieves an improvement of $0.52\%$, $0.89\%$ and $0.72\%$ compared to AlexNet in terms of \textbf{DP} and \textbf{OP} and \textbf{AUC}.
But the speed decreases from $18.8$ \textbf{FPS} to $4.1$ \textbf{FPS}.
Considering the performance gap between different CNN features is less than $2\%$ in all numerical metrics, we argue that AlexNet provides ideal deep CNN representation in order to achieve a trade-off between accuracy and efficiency. 

\subsection{Comparison with the State-of-the-art}
\subsubsection{Overall Performance}
The experimental results obtained on OTB2013, TB50 and OTB2015 are reported in Table~\ref{otb} and Fig.~\ref{otbfig}.
In Table~\ref{otb}, we present the \textbf{OP} and \textbf{CLE} on each dataset.
Our ACFT method is the best in terms of \textbf{OP} and a respectable performance in terms of \textbf{CLE}.
Compared with C-COT, our ACFT achieves an improvement of $5.1\%$, $2.9\%$ and $4.0\%$ in the mean overlap success.
Note that C-COT, MetaT, MCPF and CREST employ the family of VGG networks as their feature extractors, which are more discriminative than AlexNet that is used in our ACFT method.
In spite of that, the R\_A-ADMM optimisation enables a simultaneous increase in acceleration and accuracy.
The \textbf{FPS} of ACFT is 18.8, which outperforms other on-line learning DCF trackers with deep representation.
Similar conclusions can be reached from Fig.~\ref{otbfig}. Our ACFT ranks top in terms of \textbf{AUC} and \textbf{DP} in all three benchmarks.
For \textbf{DP}, MCPF maintains its second ranking on OTB2013, with $0.1\%$ lower than ACFT.
Moreover, ACFT achieves $90.5\%$ and $90.7\%$ of \textbf{DP} on TB50 and OTB2015 respectively, outperforming all the other trackers.

For VOT2017, we report the Baseline and the Real-time~\cite{Kristan2017a} results in Table~\ref{vot}. 
In the Baseline metric, ACFT achieves a comparable performance  to the top trackers, with $0.317$, $0.522$ and $0.238$ in terms of \textbf{EAO}, \textbf{Accuracy} and \textbf{Robustness}, respectively.
As ACFT is designed to balance the tracking performance  efficiency and accuracy, it outperforms the state-of-the-art trackers in Real-time metric.
Compared to the second best in Real-time metric, CSRDCF++ (\textbf{EAO}, \textbf{Robustness}) and SiamFC (\textbf{Accuracy}), ACFT improves the performance from 0.212 (\textbf{EAO}), 0.502 (\textbf{Accuracy}) and 0.398 \textbf{Robustness}) to 0.267, 0.517 and 0.291, respectively.

The experimental results on standard benchmarks demonstrate that ACFT achieves very impressive overall performance in both accuracy and speed.

\subsubsection{Attribute Performance}
We follow the OTB benchmark~\cite{Wu2015Object} to further analyse the tracking performance with challenging videos annotated by eleven attributes~\textit{i.e.}, background clutter (BC), deformation (DEF), fast motion (FM), in-plane rotation (IPR), low resolution (LR),illumination variation (IV), out-of-plane rotation (OPR), motion blur (MB), occlusion (OCC), out-of-view (OV), and scale variation (SV), in terms of success plots on OTB2015~\cite{Wu2015Object} in Fig.~\ref{attributes}. 
We observe that our ACFT method outperforms all the other trackers in eight attributes,~\textit{i.e.},
LR, OV, IPR, FM, DEF, SV, OPR and IV in terms of \textbf{AUC}. 
Our R\_A-ADMM optimised DCF formulation promotes adaptive temporal smoothness and robustness, focusing on the relevant appearance information. 
The results of ACFT in the other three attributes are still among the top three, demonstrating the general effectiveness and robustness of our method. 
In particular, ACFT exhibits a significant performance boost ($4.4\%$, $2.1\%$, $1.7\%$ and $2.7\%$ in terms of \textbf{AUC} as compared with the corresponding second best algorithms in the attributes of LR, SV, DEF and OPR, respectively.

\begin{figure}[t]
\begin{center}
\includegraphics[width=1\linewidth]{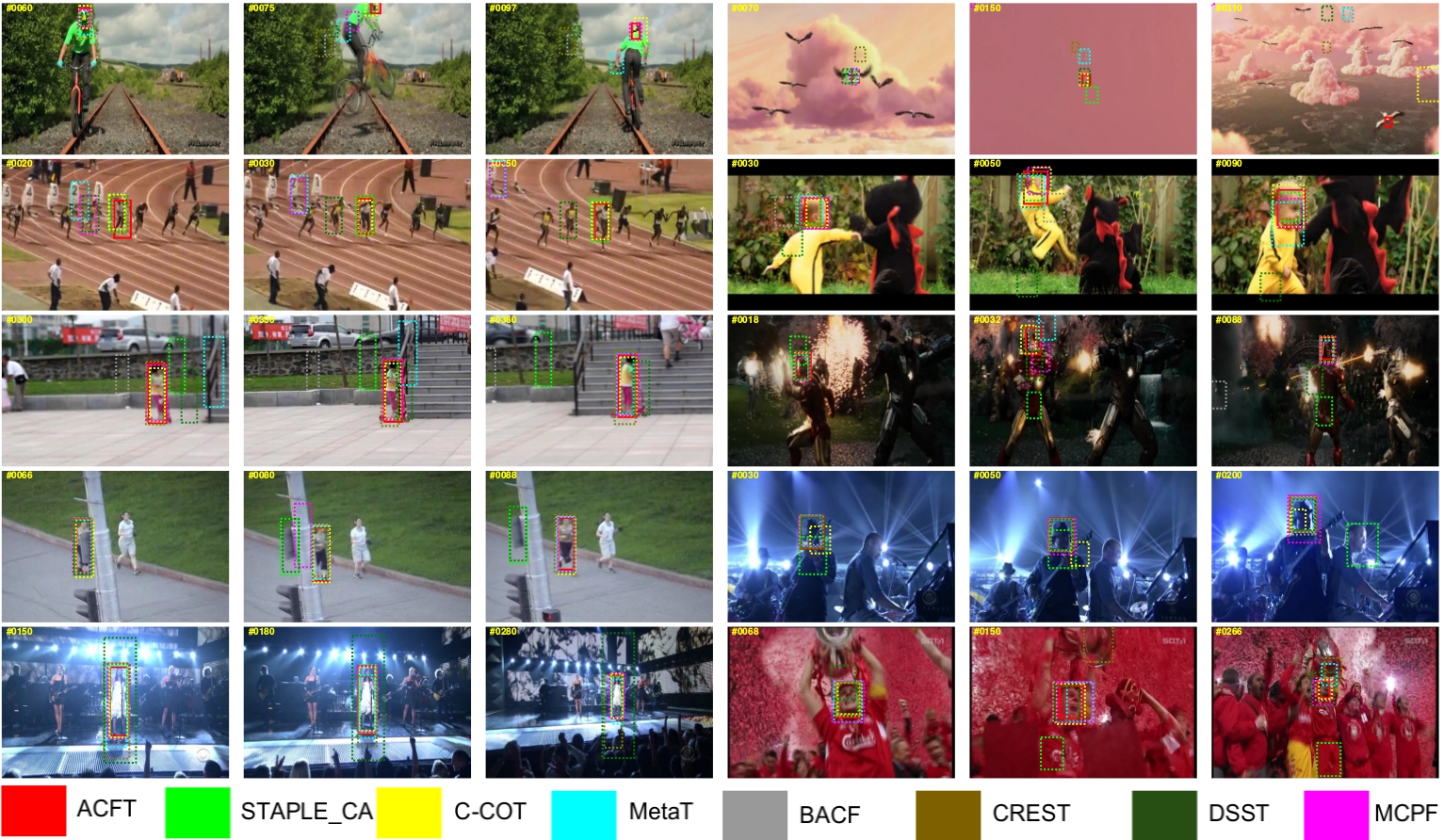}
\end{center}
\caption{A qualitative comparison of our ACFT with state-of-the-art trackers on challenging sequences of OTB100~\cite{Wu2015Object} (left column: \textit{Biker}, \textit{Bolt2}, \textit{Girl2}, \textit{Jogging-1}, and \textit{Singer1}; right column: \textit{Bird1}, \textit{Dragonboy}, \textit{Ironman}, \textit{Shaking}, and \textit{Soccer}.}\label{qualitative}
\end{figure}

Figure~\ref{qualitative} presents a qualitative comparision of our ACFT with the state-of-the-art methods,~\textit{i.e.}, STAPLE\_CA, C-COT, MetaT, BACF, CREST, DSST and MCPF, on some challenging vodep sequences.
For example, rapid appearance variations of targets and the surroundings pose severe difficulties.
ACFT performs well on these challenges, benefiting from learning the regularised DCF framework with effective optimisation. 
Sequences with deformations (\textit{Bolt2}, \textit{Dragonbaby}, and \textit{Soccer}) and out of view (\textit{Biker} and \textit{Bird1}) can  successfully be tracked by our methods without any failures. 
Videos with occlusions (\textit{Jogging-1}, \textit{Girl2}, and \textit{Bird1}) also benefit from our ACFT formulation. 
Specifically, ACFT is expert in tracking noise-corrupted targets (\textit{Bird1}, \textit{Dragonbaby}, and \textit{Soccer}), because the proposed adaptive temporal initialisation promotes global robustness across consecutive frames. 

\subsubsection{Discussions}
\myrevisedcolor{Based on the above experimental results, our ACFT achieves outstanding efficiency among deep feature based DCF trackers. The acceleration of ACFT is achieved by the combined impact of R\_A-ADMM, adaptive initialisation and the use of light-weight deep networks. 
Though light-weight deep networks, \textit{e.g.}, AlexNet, provide semantic descriptors to a certain extent, the tracking accuracy can be further improved with more powerful deep networks, \textit{e.g.}, ResNet. 
To this end, we plan, in future, to explore the techniques, such as network compression, to achieve further performance improvements.}

\section{Conclusion}\label{conclusion}
To achieve simultaneous acceleration of the process of feature extraction and on-line appearance model learning, we proposed an efficient optimisation framework for the regularised paradigm of discriminative correlation filter design.
The performance improvement has been achieved by reformulating the appearance model optimisation so as to incorporate relaxation and acceleration, with additional adaptive initialisation of the filter learning process for enhanced temporal smoothness. 
Light-weight deep features are employed to balance the tracking performance between accuracy and efficiency.
The results of extensive experimental studies performed on tracking benchmark datasets demonstrated the effectiveness and robustness of our method, compared to the state-of-the-art trackers.

\section*{Acknowledgment}
This work was supported in part by the UK Engineering and Physical Sciences Research Council (EPSRC) under grant number EP/R013616/1, the EPSRC Programme Grant (FACER2VM) EP/N007743/1, the National Natural Science Foundation of China (61672265, U1836218, 61876072, 61902153) and the 111 Project of Ministry of Education of China (B12018).

\appendix
\section{The continuous limit of Algorithm~\ref{alg}.}
\label{FirstAppendix}
To establish a connection between the iterative optimisation specified in  Algorithm~\ref{alg} and its  continuous dynamical system equivalent, we rearrange the variable expressions in Equ.~(\ref{aug}) and Equ.~(\ref{aug_d}) as:
\begin{equation}
\label{augp}
\mathcal{L}_{\rho}\left(\hat{\mathbf{u}},\mathbf{w},\mathbf{\tau}\right)=f\left(\hat{\mathbf{u}}\right)+g\left(\mathbf{w}\right)+h_{\rho}\left(\hat{\mathbf{u}},\mathbf{w},\mathbf{\tau}\right),
\end{equation}
\begin{equation}\label{aug_dp}
\left\{
\begin{aligned}
&f\left(\hat{\mathbf{u}}\right)=\left\|\hat{\mathbf{x}}\odot\hat{\mathbf{u}}-\hat{\mathbf{y}}\right\|_2^2\\
&g\left(\mathbf{w}\right)=\lambda_1\left\|\mathbf{w}\odot\mathbf{p}\right\|_2^2\\
&h_{\rho}\left(\hat{\mathbf{u}},\mathbf{w},\mathbf{\tau}\right)=\frac{\rho}{2}\left\|\mathcal{F}^{-1}\left(\hat{\mathbf{u}}\right)-\mathbf{w}+\mathbf{\tau}\right\|_2^2\\
\end{aligned}
\right.,
\end{equation}
where $\mathbf{u}$, $\mathbf{w}$, and $\mathbf{\tau}$ are the vectorised forms of $\mathcal{U}$, $\mathcal{W}$, $\mathcal{T}$. $\mathbf{y}$ and $\mathbf{p}$ are the vectorised forms of $\mathbf{Y}$ and $\mathbf{P}$ duplicated $C$ times.
The Fourier Transform of a vector can be considered as a linear mapping by the DFT matrix $\mathbf{F}$:
\begin{equation}
    \mathcal{F}\left(\mathbf{w}\right)=\mathbf{F}\mathbf{w},\ \ \ \ \ \mathcal{F}^{-1}\left(\hat{\mathbf{u}}\right)=\mathbf{F}^H\hat{\mathbf{u}}
\end{equation}
$\mathbf{F}$ is a unitary matrix ($\mathbf{F}^H\mathbf{F}=\mathbf{I}$), reflecting the natural transform between $\hat{\mathbf{u}}$ and $\mathbf{w}$, $(\cdot)^H$ denotes Hermitian transpose. 
We obtain the following equations via the optimality conditions of Step 1 and Step 3 of Algorithm~\ref{alg}
\begin{equation}
\nabla　f\left(\hat{\mathbf{u}}[l+1]\right)+\rho\mathbf{F}\left(\mathbf{F}^H\hat{\mathbf{u}}[l+1]-\mathbf{w}^{\prime}[l]+\mathbf{\tau}^{\prime}[l]\right)=0,
\end{equation}
\begin{equation}
\begin{aligned}
&\nabla g\left(\mathbf{w}[l+1]\right)+\rho\mathbf{F}\left\{\mathbf{F}^H\left(\alpha\hat{\mathbf{u}}[l+1]+\left(1-\alpha\right)\mathbf{F}\mathbf{w}^\prime[l]\right)\right.\\
&\left.-\mathbf{w}[l+1]+\mathbf{\tau}^{\prime}[l]\right\}=0.
\end{aligned}
\end{equation}
Such that:
\begin{equation}\label{d1}
\begin{aligned}
&\nabla f\left(\hat{\mathbf{u}}[l+1]\right)+\mathbf{F}\nabla g\left(\mathbf{w}[l+1]\right)+\rho\left(1-\alpha\right)\mathbf{F}\\
&\times(\mathbf{F}^H\hat{\mathbf{u}}[l+1]-\mathbf{w}^\prime[l])+\rho\mathbf{F}\left(\mathbf{w}[l+1]-\mathbf{w}^\prime[l]\right)=0.
\end{aligned}
\end{equation}
From Step 7 in Algorithm~\ref{alg}, we get:
\begin{equation}
\begin{aligned}
&\mathbf{w}[l+1]-\mathbf{w}^\prime[l]=(\mathbf{w}[l+1]-2\mathbf{w}[l]+\mathbf{w}[l-1])\\
&+(1-\beta)(\mathbf{w}[l]-\mathbf{w}[l-1]).
\end{aligned}
\end{equation}
Let $\hat{\mathbf{u}}[l] = \hat{\mathbf{U}}(t), \mathbf{w}[l]=\mathbf{W}(t), \mathbf{\tau}[l]=\mathbf{T}(t), \mathbf{w}^\prime[l]=\mathbf{W}^\prime(t), \mathbf{\tau}^{\prime}[l]=\mathbf{T}^{\prime}(t)$ with $t=l\epsilon$.
Setting $\rho=1/\epsilon^2$, it is intuitive that $\rho(\mathbf{w}[l+1]-2\mathbf{w}[l]+\mathbf{w}[l-1])\rightarrow\ddot{\mathbf{W}}(t)$ and $(1-\beta)(\mathbf{w}[l]-\mathbf{w}[l-1])\rightarrow\frac{r}{t}\dot{\mathbf{W}}(t)$ as $\epsilon\rightarrow 0$.
Therefore, we obtain $\rho(\mathbf{w}[l+1]-\mathbf{w}^\prime[l])\rightarrow\ddot{\mathbf{W}}(t)+\frac{r}{t}\dot{\mathbf{W}}(t)$.
Combined with Step 4 in Algorithm~\ref{alg}, we get $\mathbf{F}^H\hat{\mathbf{U}}(t)=\mathbf{W}(t)$, $\mathbf{F}^H\dot{\hat{\mathbf{U}}}(t)=\dot{\mathbf{W}}(t)$ and $\mathbf{F}^H\ddot{\hat{\mathbf{U}}}(t)=\ddot{\mathbf{W}}(t)$, such that:
\begin{equation}
\rho\left(\mathbf{F}^H\hat{\mathbf{u}}[l+1]-\mathbf{w}^\prime[l]\right)\rightarrow\ddot{\mathbf{W}}(t)+\frac{r}{t}\dot{\mathbf{W}}(t).
\end{equation}
Substituting the corresponding variables into Equ.~(\ref{d1}), we obtain:
\begin{equation}
\begin{aligned}
&\nabla f\left(\hat{\mathbf{U}}(t)\right)+\mathbf{F}\nabla g\left(\mathbf{W}(t)\right)\\
&+(2-\alpha)\mathbf{F}^H\mathbf{F}\left(\ddot{\mathbf{W}}(t)+\frac{r}{t}\dot{\mathbf{W}}(t)\right)=0.
\end{aligned}
\end{equation}
Following the definition of $\Lambda\left(\right)$ in Equ.~(\ref{objre}), we obtain the final equivalent form:
\begin{equation}
(2-\alpha)\left(\ddot{\mathbf{W}}(t)+\frac{r}{t}\dot{\mathbf{W}}(t)\right)+\nabla\Lambda\left(\mathbf{W}(t)\right)=0.
\end{equation}

\section*{References}
\bibliography{ref.bib}
\end{document}